\documentclass[electronic]{vgtc}             % electronic version

\ifpdf%                                % if we use pdflatex
  \pdfoutput=1\relax                   % create PDFs from pdfLaTeX
  \usepackage{graphicx}                % allow us to embed graphics files
  \DeclareGraphicsExtensions{.pdf,.png,.jpg,.jpeg} % for pdflatex we expect .pdf, .png, or .jpg files
\else%                                 % else we use pure latex
  \usepackage{graphicx}                % allow us to embed graphics files
  \DeclareGraphicsExtensions{.eps}     % for pure latex we expect eps files
\fi%

\graphicspath{{figures/}{pictures/}{images/}{./}} % where to search for the images

\usepackage{microtype}                 % use micro-typography (slightly more compact, better to read)
\PassOptionsToPackage{warn}{textcomp}  % to address font issues with \textrightarrow
\usepackage{textcomp}                  % use better special symbols
\usepackage{mathptmx}                  % use matching math font
\usepackage{times}                     % we use Times as the main font
\usepackage{cite}                      % needed to automatically sort the references
\usepackage{tabu}                      % only used for the table example
\usepackage{booktabs}                  % only used for the table example
\usepackage{amsmath}
\usepackage{comment}
\usepackage[table,xcdraw]{xcolor}
\usepackage{multirow}

\newcommand{\blue}{\textcolor{blue}}

\onlineid{2111}

\vgtccategory{Research}

\vgtcinsertpkg

\title{PoP-Net: Pose over Parts Network for Multi-Person 3D Pose Estimation from a Depth Image}

\author{Yuliang Guo\thanks{Contact Author, the work was done when Guo was with OPPO.}\\ 
\and Zhong Li\thanks{Contact Author.}\\
\and Zekun Li\thanks{The work was done when Li was an intern with OPPO.}\\     
\and Xiangyu Du\\ 
\and Shuxue Quan\\
\and Yi Xu
}
\affiliation{\scriptsize \{yuliang.guo, zhong.li, zekun.li, xiangyu.du, shuxue.quan, yi.xu\}@oppo.com\\ OPPO US Research Center}
\abstract{In this paper, a real-time method called PoP-Net is proposed to predict multi-person 3D poses from a depth image. PoP-Net learns to predict bottom-up part representations and top-down global poses in a single shot. Specifically, a new part-level representation, called Truncated Part Displacement Field (TPDF), is introduced which enables an explicit fusion process to unify the advantages of bottom-up part detection and global pose detection. Meanwhile, an effective mode selection scheme is introduced to automatically resolve the conflicting cases between global pose and part detections. Finally, due to the lack of high-quality depth datasets for developing multi-person 3D pose estimation, we introduce Multi-Person 3D Human Pose Dataset (MP-3DHP) as a new benchmark. MP-3DHP is designed to enable effective multi-person and background data augmentation in model training, and to evaluate 3D human pose estimators under uncontrolled multi-person scenarios. We show that PoP-Net achieves the state-of-the-art results both on MP-3DHP and on the widely used ITOP dataset, and has significant advantages in efficiency for multi-person processing. To demonstrate one of the applications of our algorithm pipeline,  we also show results of virtual avatars driven by our calculated 3D joint positions. MP-3DHP Dataset and the evaluation code 
have been made available at: \url{https://github.com/oppo-us-research/PoP-Net}.
} % end of abstract

\begin{document}

\maketitle

\section{Introduction}

Human pose estimation plays an important role in a wide variety of applications, and there is a rich pool of literature for human pose estimation methods. Categorizations of existing methods can be made from different dimensions. There are methods mostly relying on a single image to predict human poses~\cite{Wei:etal:CVPR2016:CPM,DBLP:conf/cvpr/PavlakosZDD17a,Mehta:etal:ACM2017:VNect} and others based on multiple cameras~\cite{DBLP:conf/eccv/RhodinRCRST16,Elhayek:etal:PAMI17}. Some methods are capable of predicting multiple poses~\cite{Cao:etal:cvpr17:openpose,Mehta:etal:XNECT:TOG2020} while others are focusing on single person~\cite{Wei:etal:CVPR2016:CPM,Newell:etal:ECCV2016:hourglass,DBLP:conf/cvpr/PavlakosZDD17a}. Some methods estimate 3D poses~\cite{DBLP:conf/bmvc/TekinKSLF16,DBLP:conf/iccv/MartinezHRL17,DBLP:conf/cvpr/PavlakosZD18,Xiong:etal:iccv2019:a2j} while others only predict 2D poses~\cite{Wei:etal:CVPR2016:CPM,Newell:etal:ECCV2016:hourglass,DBLP:conf/cvpr/PapandreouZKTTB17}. Methods can also be differentiated by the input, while most methods use RGB images~\cite{Cao:etal:cvpr17:openpose,DBLP:conf/cvpr/PapandreouZKTTB17,Mehta:etal:ACM2017:VNect,BenzineCLPA:CVPR2020}, some use depth maps~\cite{Martiez-Gonzalez:etal:iros2018:rpm,Wang:etal:ACM2016,Xiong:etal:iccv2019:a2j}. Specifically, this paper focuses on multi-person 3D pose estimation from a depth image.
% We consider to provide a unified network capable of solving these two tasks because recovering the true depth from an noise depth map is relative simple without much sophisticated design compared to the task of extracting 3D pose from a single RGB image. 

%\begin{figure}[!h]
%    \centering
%    \includegraphics[width=0.49\textwidth]{figs/pipeline_summary.pdf}
%    \caption{ \textbf{PoP-Net paradigm.} PoP-Net is end-to-end learned to predict three part-level representations and an anchor-based global pose map. In the fuse process, the proposed Truncated-Part-Displacement-Field (TPDF) maps are utilized to drag predicted 3D global poses towards more accurate bottom-up part predictions. Meanwhile, part confidence maps and part depths maps are leveraged to estimate more reliable 3D poses via weighted aggregation.}
%  \label{fig:our:pipeline}
%\end{figure}

\begin{figure}[!t]
    \centering
    \includegraphics[width=0.47\textwidth]{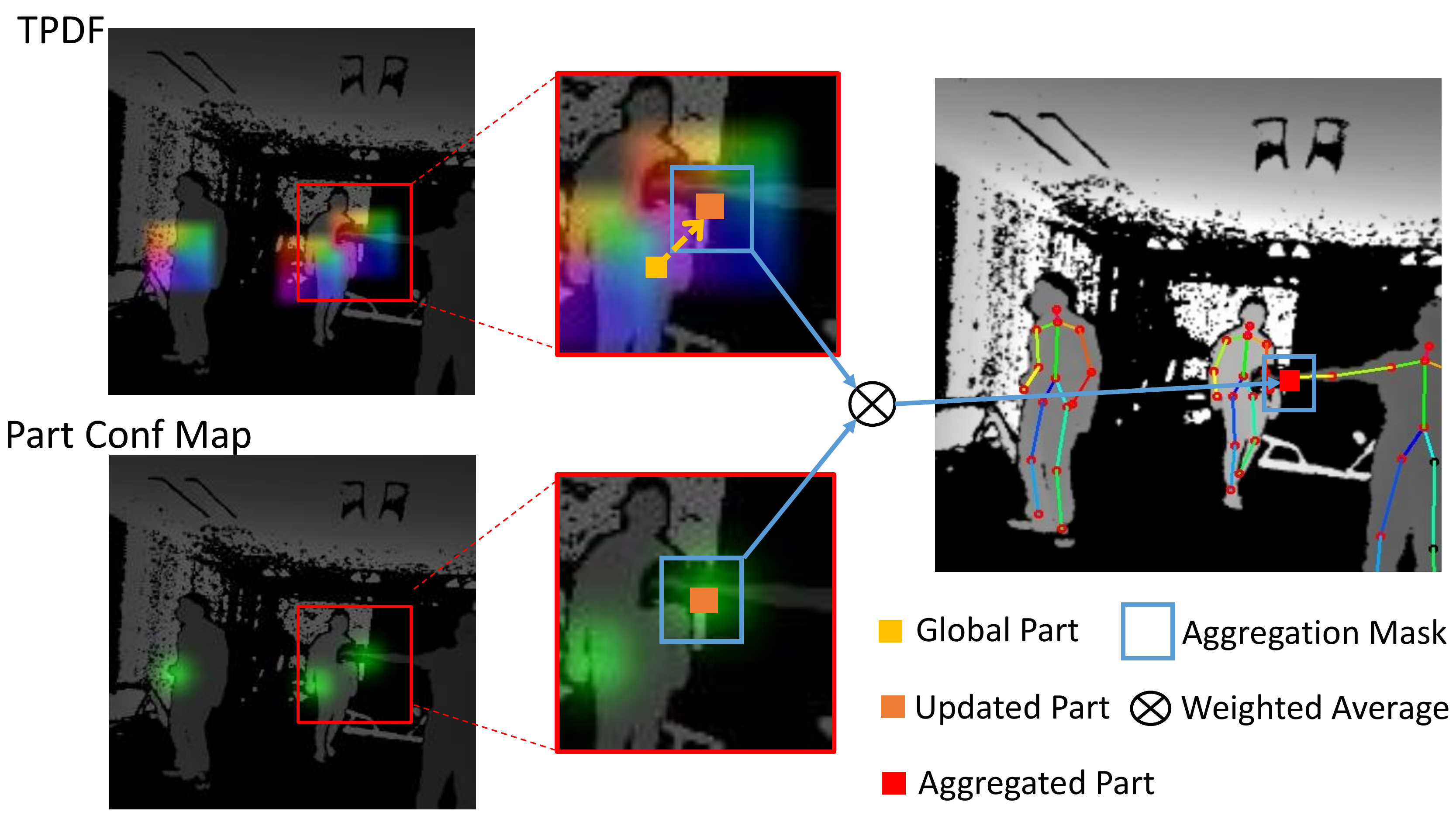}
    \caption{ \textbf{Our paradigm.} Part representations and global poses predicted from PoP-Net are explicitly fused via utilizing Truncated-Part-Displacement-Field (TPDF). A part predicted from the global pose is dragged towards a more precise bottom-up part position following a displacement vector. More reliable part position is further estimated via a part-confidence-weighted average of TPDF within the aggregation mask.}
  \label{fig:main:idea}
\end{figure}

%From the perspective of methodology, multi-person pose estimation task has been approached from both classic trend and DNN-based trend. A classic method normally formulates the problem as maximizing a posterior given a part probabilistic model of both the appearance of the parts and pairwise spatial constraints between the parts. Approaches along the classic trend include~\cite{Felzenszwalb:etal:PAMI2010:DPM,Yang:etal:PAMI2013:FMP} and their variants. 
%In the era of deep learning, Deep Neural Networks (DNN)-based methods~\cite{Wei:etal:CVPR2016:CPM,Newell:etal:ECCV2016:hourglass,Cao:etal:cvpr17:openpose,He:etal:iccv2017:mask:rcnn,Mehta:etal:XNECT:TOG2020,Xiong:etal:iccv2019:a2j} have been developed. %The foundation of DNN-based network relies on large amount of available pose data such that independent part detection is robust enough without the necessity to explicitly modeling the pairwise constraints. Those pair-wise constraints can be interpreted as being implicitly encoded in the appearance context of each part. State-of-the-art methods reported on the most popular human pose datasets (e.g., COCO~\cite{xx}, ITOP~\cite{Haque:etal:eccv2016:itop}) are all DNN-based.

\begin{figure*}[!h]
    \centering
    \includegraphics[width=0.17\textwidth]{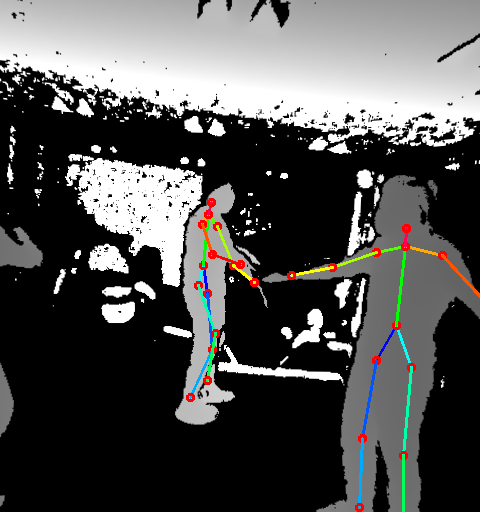}
    \includegraphics[width=0.17\textwidth]{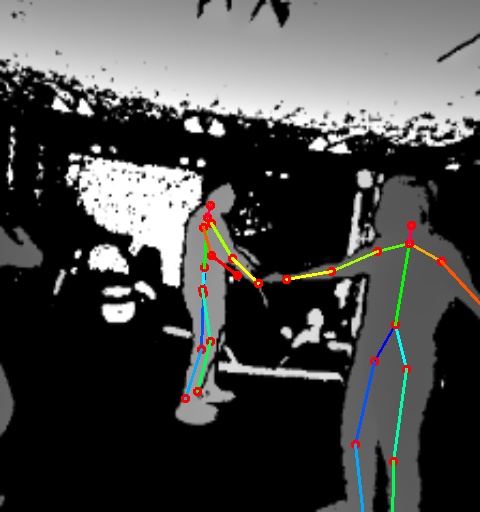}
    \includegraphics[width=0.17\textwidth]{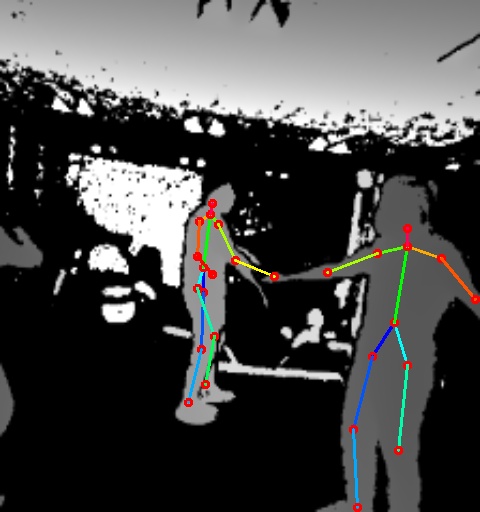}
    \includegraphics[width=0.17\textwidth]{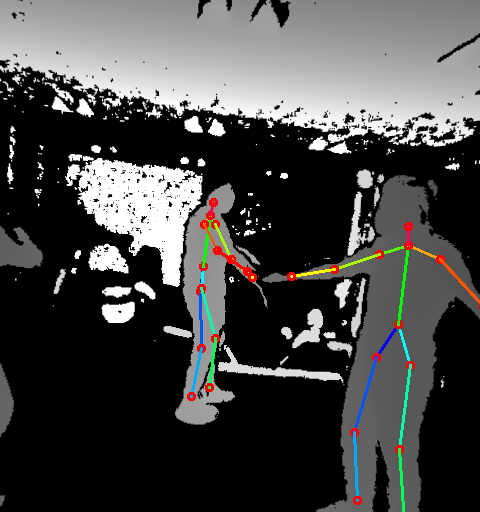}
    \includegraphics[width=0.17\textwidth]{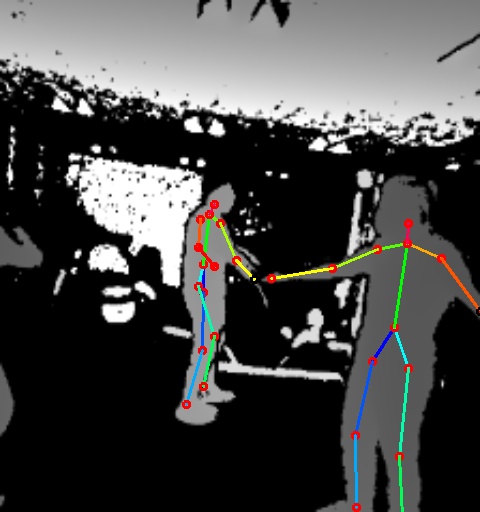} \\
    \includegraphics[width=0.17\textwidth]{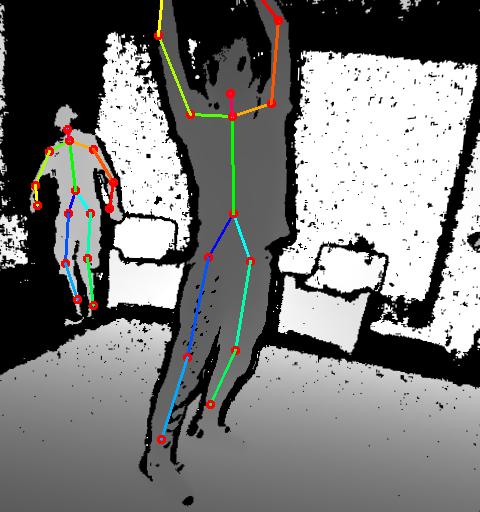}
    \includegraphics[width=0.17\textwidth]{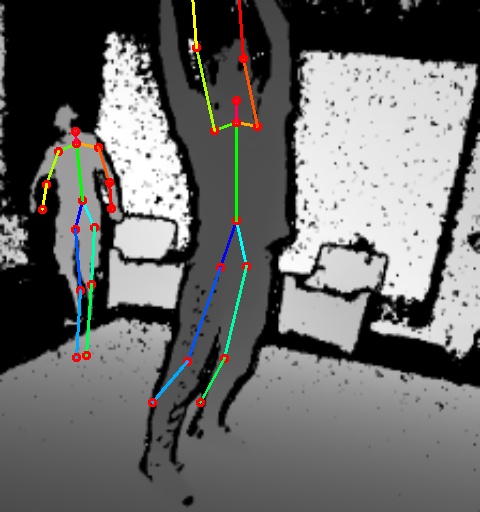}
    \includegraphics[width=0.17\textwidth]{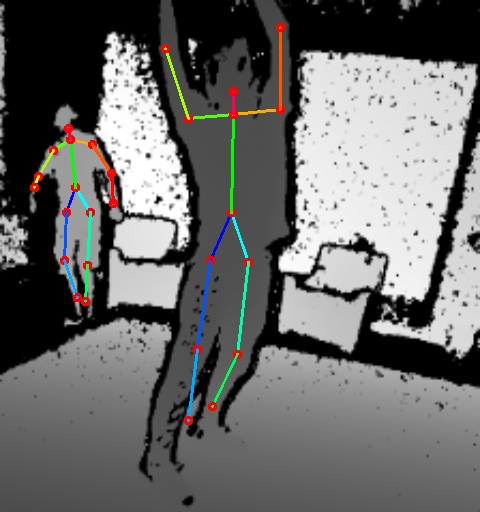}
    \includegraphics[width=0.17\textwidth]{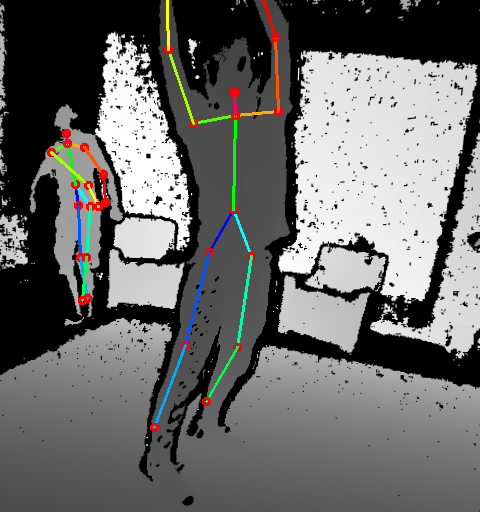}
    \includegraphics[width=0.17\textwidth]{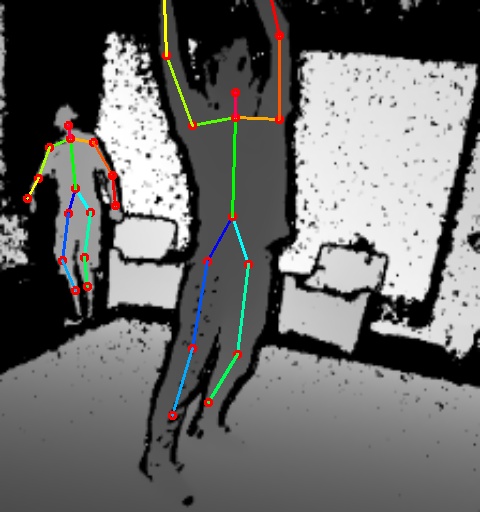}\\
    Ground-Truth \qquad\qquad Yolo-Pose+ \qquad\qquad\quad Open-Pose+ \qquad\qquad Yolo-A2J \qquad\qquad\quad PoP-Net \quad
\caption{ \textbf{Visual comparison of prototypical methods:} methods are compared on two examples from MP-3DHP testing set.}
 \label{fig:visual:compare}
\end{figure*}

%~\cite{Wei:etal:CVPR2016:CPM,Newell:etal:ECCV2016:hourglass,Cao:etal:cvpr17:openpose,He:etal:iccv2017:mask:rcnn,DBLP:conf/cvpr/PapandreouZKTTB17,Xiong:etal:iccv2019:a2j}
In the era of deep learning, a large pool of Deep Neural Networks (DNN)-based methods have been developed for multi-person pose estimation. Ideas from existing literature can be generally categorized into three prototypical trends. The simplest idea is to directly extend a single-shot object detector~\cite{Liu:etal:eccv2016:ssd,Redmon:etal:cvpr16:yolo,Redmon:Farhadi:cvpr2017:yolo2} with additional pose attributes, so that the network can output human poses. Such single-shot regression can be very efficient, but has low part accuracy, as shown in Figure~\ref{fig:visual:compare} (Yolo-Pose+), because a long-range inference for part locations is involved in a center-relative pose representation. The second type builds on a two-stage pipeline where the first stage detects object bounding boxes, and the second stage estimates the pose within each~\cite{DBLP:conf/cvpr/PapandreouZKTTB17,He:etal:iccv2017:mask:rcnn,Xiong:etal:iccv2019:a2j}. Two-stage methods can be very accurate, as shown in Figure~\ref{fig:visual:compare} (Yolo-A2J), but not as efficient when more human beings appear in an image. In addition, more sophisticated design is required to solve the compatibility issue between pose estimation and bounding box detection~\cite{He:etal:iccv2017:mask:rcnn,Redmon:etal:cvpr16:yolo}. The third idea is to detect human poses from part\footnote{The definitions of 'part' and 'joint' are interchangeable in this paper.} detection and  association~\cite{DBLP:conf/eccv/IqbalG16,DBLP:conf/nips/NewellHD17,Cao:etal:cvpr17:openpose,Martiez-Gonzalez:etal:iros2018:rpm,Mehta:etal:XNECT:TOG2020}. Although part detection can be rather efficient, solving the part association problem is usually time consuming. OpenPose~\cite{Cao:etal:cvpr17:openpose} gained its popularity for introducing an efficient solution to solve the association, resulting in a network benefiting from both the single-shot pipeline with high efficiency and the part-based dense representation with high positional precision. However, a pure bottom-up method does not have a global sense, so that it is rather sensitive to occlusion, truncation, and ambiguities in symmetric limbs (Figure~\ref{fig:visual:compare} OpenPose+). Moreover, dependency on the bipartite matching in assembling parts presents the part predictions from differentiable integration with global context reasoning, which may systematically block a solution from end-to-end training~\cite{Mehta:etal:XNECT:TOG2020}.

%\begin{figure}[!h]
%    \centering
%    \includegraphics[width=0.48\textwidth]{figs/visual_compare.pdf}
%    \caption{ {\bf Visual comparison of prototypical methods:} Prototypical methods are visually compared on three examples from the multi-person testing set. From the top row to the bottom, we show ground-truth poses, yolo-pose output, Open-Pose+ output and our method pop-net output.}
%  \label{fig:visual:compare}
%\end{figure}

Developing a depth image-based multi-person 3D pose estimation from a well-established RGB image-based 2D method~\cite{DBLP:conf/cvpr/PapandreouZKTTB17,Ren:etal:pami2017:faster:rcnn,Cao:etal:cvpr17:openpose} is conceptually simple for two reasons: some schemes to handle multi-person detection can be shared; 3D information is partially available from the input~\cite{Haque:etal:eccv2016:itop,Xiong:etal:iccv2019:a2j}. As a result, a depth image-based approach may not necessarily require as many sophisticated designs as methods aiming to estimate 3D poses from a single RGB image~\cite{Mehta:etal:ACM2017:VNect,Mehta:etal:XNECT:TOG2020}. However, RGB-based methods and depth-based methods usually focus on different challenges. While some RGB-based methods spend more effort in differentiating people cluttered in a large scene~\cite{Cao:etal:cvpr17:openpose,BenzineCLPA:CVPR2020}, depth-based methods~\cite{Martiez-Gonzalez:etal:iros2018:rpm,Xiong:etal:iccv2019:a2j} focus more on recovering highly accurate 3D poses of fewer people presented in a closer range. In practice, to develop a depth-based method that robustly predicts 3D poses in a multi-person scenario is very challenging due to noisy depth inputs and heavy occlusions. %Consequently, the design of our method focuses on estimating highly accurate close/middle-range 3D poses under occlusion and noisy inputs.

% the majority of the effort has beenspent on delivering accurate estimation of multiple 2D poses and the fusion of available depth information from different modules.\zhong{The literature of Depth-based 3D human pose estimation methods have two directions, use generative and discriminate models. The generative models use ICP alogrithm to estimate the human pose by find correspondence between predefined body template and input human body point cloud~\cite{xxxx}, those method is inaccurate and is sensitive to the input noise. Discriminate models don't require body template and direct estimate the position of body parts based on random forest, well known work Shotton et al.~\cite{shotton2011real} use self-defined feature to classify each pixel into one of the body parts, while Juny et al.~\cite{yub2015random} directly regress the coordniate of the body part using a random tree walk algorithm. Recently, several 2D and 3D CNN based method is proposed, Haque et al.~\cite{Haque:etal:eccv2016:itop} proposed a CNN solution that could learn view-invariant feature hence the method is robust to varies challenging viewpoints. Moon et al. ~\cite{V2V} proposed a 3d voxel-to-vodel prediction network~\cite{A2J}, however 3D CNN is not cost efficient in terms of running time and memory storage. Xiong et al.~\cite{Xiong:etal:iccv2019:a2j} develop a much faster 3D supervised method based on 2d offset predictions and depth predictions with anchor points. Nevertheless, above CNN based methods is conduct under the assumption that the human region is known. }

In this paper, we present a method called Pose-over-Parts Network (\textbf{PoP-Net}) to estimate multiple 3D poses from a depth image. As illustrated in Figure~\ref{fig:main:idea}, the main idea of PoP-Net is to explicitly fuse the predicted bottom-up parts and top-down global poses\footnote{Poses predicted from a single-shot network where each pose contains a full set of body parts}. This fusion process is enabled by a new intermediate representation, called Truncated-Part-Displacement-Field (TPDF), which is a vector field that records the vector pointing to the closest part location at every 2D position. TPDF is utilized to guide a structural valid global pose towards more positionally precise part location so that the advantages of global pose and local part detection can be naturally unified.

At the same time, we release a comprehensive depth dataset, named Multi-Person 3D Human Pose Dataset (MP-3DHP), to facilitate the development of 3D pose estimation methods generalizable to novel background and unobserved multi-person configurations in real-world applications. Although there are a decent amount of RGB datasets~\cite{DBLP:conf/eccv/LinMBHPRDZ14,DBLP:journals/pami/IonescuPOS14,DBLP:conf/cvpr/AndrilukaPGS14:mpii} in prior art, there are limited high-quality depth datasets. The released dataset is constructed to cover most of the essential aspects of visual variance related to 3D human pose estimation, and particularly to promote the development of multi-person methods. %There is limited data in the prior art for developing DNN-based methods for poes estimation using depth images. There is indeed  sufficient data for 2D poses and RGB images(\zhong{problem of RGB 2D pose}) when the search space is limited in 2D and the domain gap between the training data and testing data is manipulable. However, due to larger domain gap between data collected from depth sensors and larger search space in 3D pose, there are limited dataset available~\cite{xxx,xxx}. ~\zhong{talk about serveral art dataset.. and their draw back}. 

The contribution of this paper is four fold. First, we introduce an efficient framework that predicts multiple 3D poses in a single shot. Second, we propose a new part-level representation called TPDF, which enables an explicit fusion of global poses and part-level representations. Third, we introduce a mode selection scheme that automatically resolves the conflicting cases between local and global predictions. Finally, we release a comprehensive depth image-based dataset to facilitate the development of multi-person 3D pose estimation methods applicable to real-world challenges.

\begin{figure*}[!t]
    \centering
    \includegraphics[width=0.85\textwidth]{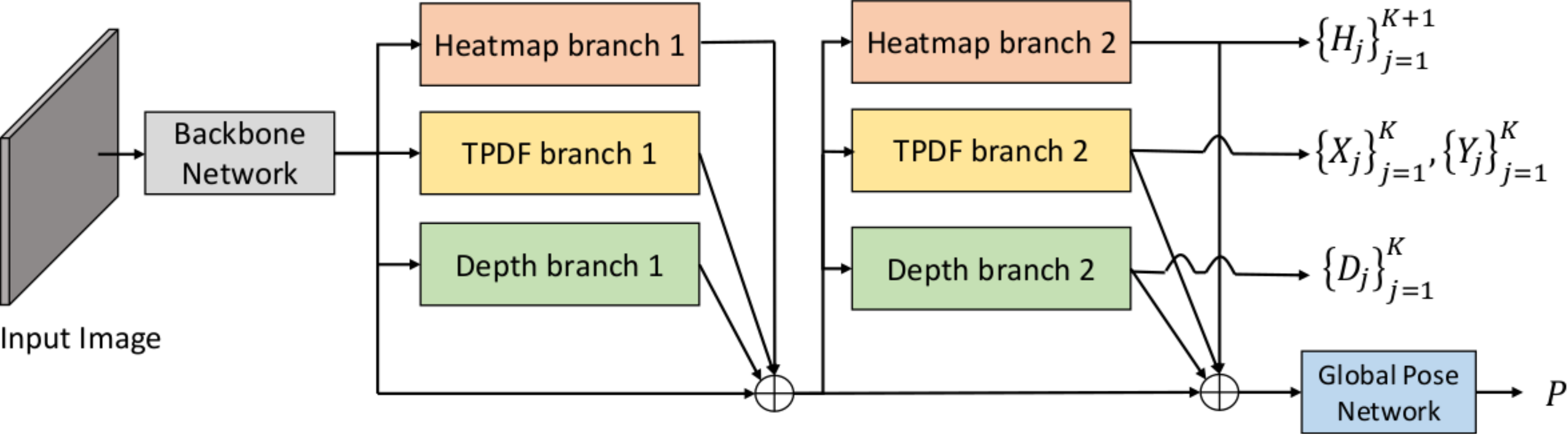}
    \caption{ \textbf{PoP-Net} is composed of a backbone network, three functional branches, and a global pose network. The functional branches are organized in two stages with split and merge. PoP-Net outputs three part-level maps and a global pose map.}
  \label{fig:net:arch}
\end{figure*}

%%%%%%%%%%%%%%%%%%%%%%%%%%%%%%%%%%%%%%%%%%%%%%%%%%%%%%%%%%%%%%%%%%%%%%%%%%%%%%%%%%%%%%%%%%%%%%%%%%
%%%%%%%%%%%%%%%%%%%%%%%%%%%%%%%%%%%%%%%%%%%%%%%%%%%%%%%%%%%%%%%%%%%%%%%%%%%%%%%%%%%%%%%%%%%%%%%%%%
%%%%%%%%%%%%%%%%%%%%%%%%%%%%%%%%%%%%%%%%%%%%%%%%%%%%%%%%%%%%%%%%%%%%%%%%%%%%%%%%%%%%%%%%%%%%%%%%%%
%%%%%%%%%%%%%%%%%%%%%%%%%%%%%%%%%%%%%%%%%%%%%%%%%%%%%%%%%%%%%%%%%%%%%%%%%%%%%%%%%%%%%%%%%%%%%%%%%%

\section{Pose-over-Parts Network}

In this paper, we present a new method, called Pose-over-Parts Network (PoP-Net), for multi-person 3D pose estimation from a depth image. Our method first uses an efficient single-shot network to predict part-level representations and global poses, and then fuses the positionally precise part detection and structurally valid global poses in an explicit way. %PoP-Net is applicable to both 3D pose estimation from a depth image and 2D pose estimation from an RGB image. We do not consider PoP-Net to solve 3D pose estimation from a single RGB image because more sophisticated design to infer 3D from 2D is required.

The pipeline of PoP-Net is composed of a backbone network, a global pose network, and three functional branches: heatmap branch, depth branch, and TPDF branch, as illustrated in Figure~\ref{fig:net:arch}. The two-stage split-and-merge design is inspired by OpenPose~\cite{Cao:etal:cvpr17:openpose} and mostly follows the simplified version applied to depth input~\cite{Martiez-Gonzalez:etal:iros2018:rpm}. PoP-Net outputs three sets of part maps from the second stage of functional branches and an anchor-based global pose map from the global pose network.

Supposing a human body includes $K$ body parts, the heatmap branch outputs a set of part confidence maps $\{H_j\}^{K+1}_{j=1}$, where each $H_j$ from the first $K$ maps indicates the confidence of a body part occurring at each discrete location, and the last indicates background confidence. The depth branch outputs a set of maps $\{D_j\}^K_{j=1}$, where each $D_j$ encodes the depth values associated with part $j$. 

The core of our method is a new part-level representation called Truncated Part Displacement Field (TPDF). For each part type $j$, TPDF records a displacement vector pointing to the \textit{closest} part instance at every 2D position. The TPDF branch outputs TPDFs represented in a set of \textit{x}-axis displacement maps $\{X_j\}^{K}_{j=1}$ and a set of \textit{y}-axis displacement maps $\{Y_j\}^{K}_{j=1}$. The novelty of the proposed TPDF is two fold: (1) it encodes the displacement field involving multiple parts of the same type in a single map, and (2) a \textit{truncated} effective range is adopted, which is critical for training CNN models that are able to handle multi-body scenarios. If truncated range is not applied to the part displacement field involving part instances from multiple bodies, the training of convolutional kernels will be confused by image patches similar in appearance but associated with highly different vectors, because a pair of displacement vectors whose origins are close to each other but pointing to different part instances may have 
large difference in $X, Y$ values. The effectiveness of applying the truncated range is analyzed in detail in Section~\ref{sec:detail:ablation}.

Compared with previous methods which predict person-wise part displacements~\cite{DBLP:conf/cvpr/PapandreouZKTTB17,Xiong:etal:iccv2019:a2j}, TPDF operates at image level. In consequence, PoP-Net not only handles multiple bodies in one pass but also happens to be less sensitive to proposal error. While compared with the Part Affinity Field introduced in OpenPose~\cite{Cao:etal:cvpr17:openpose}, TPDF is free from the heavy bipartite matching process, and uses a simple fusion process to take advantage of both global poses and bottom-up part detections. As a result, PoP-Net shows increased robustness to handling truncation, occlusion and multi-person conflict compared with OpenPose.

Finally, a global pose map $P$ is regressed from the global pose network. The global pose network is a direct extension from Yolo2~\cite{Redmon:Farhadi:cvpr2017:yolo2}, where both bounding box attributes and additional 3D pose attributes are regressed with respect to the anchors associated with each grid. A set of predicted global poses are then extracted via conducting NMS on the global pose map $P$.

\subsection{Training}

PoP-Net is trained end-to-end via minimizing the total loss $\pounds$ which is the sum of heatmap loss $\pounds_{h}$, depth loss $\pounds_{d}$, TPDF loss $\pounds_{t}$, and global pose loss $\pounds_{p}$. As shown in Figure~\ref{fig:net:arch}, losses corresponding to the functional branches are contributed from multiple stages of the network. Specifically, the loss function can be written as:
\begin{align*}
    \pounds &= \pounds_{h} + \pounds_{d} + \pounds_{t} + \pounds_{p}\\
    \pounds_{h} &= \sum^S_{s=1} \sum^{K+1}_{j=1} \left\| H^s_j - H^*_j \right\|^2_2\\
    \pounds_{d} &= \sum^S_{s=1} \sum^{K}_{j=1} W^{d}_j \cdot \left\| D^s_j - D^*_j \right\|^2_2\\
    \pounds_{t} &= \sum^S_{s=1} \sum^{K}_{j=1} W^{t}_j \cdot (\left\| X^s_j - X^*_j \right\|^2_2 + \left\| Y^s_j - Y^*_j \right\|^2_2) \\
    \pounds_{p} &= W^{p} \cdot \left\| P - P^* \right\|^2_2
\end{align*}
where $s$ is the stage index, and $S$ indicates the total number of stages. $H^*_j$, $D^*_j$, $X^*_j$, $Y^*_j$, and $P^*$ indicate the ground-truth maps while $W^d_j$, $W^t_j$, and $W^p$ indicate the point-wise weight maps in the same dimension as the corresponding ground-truth maps. Specifically, no weight maps are applied to heatmap loss so that the foreground and background samples are treated with equal importance. 

\begin{figure}[!h]
    \centering
    \includegraphics[width=0.45\textwidth]{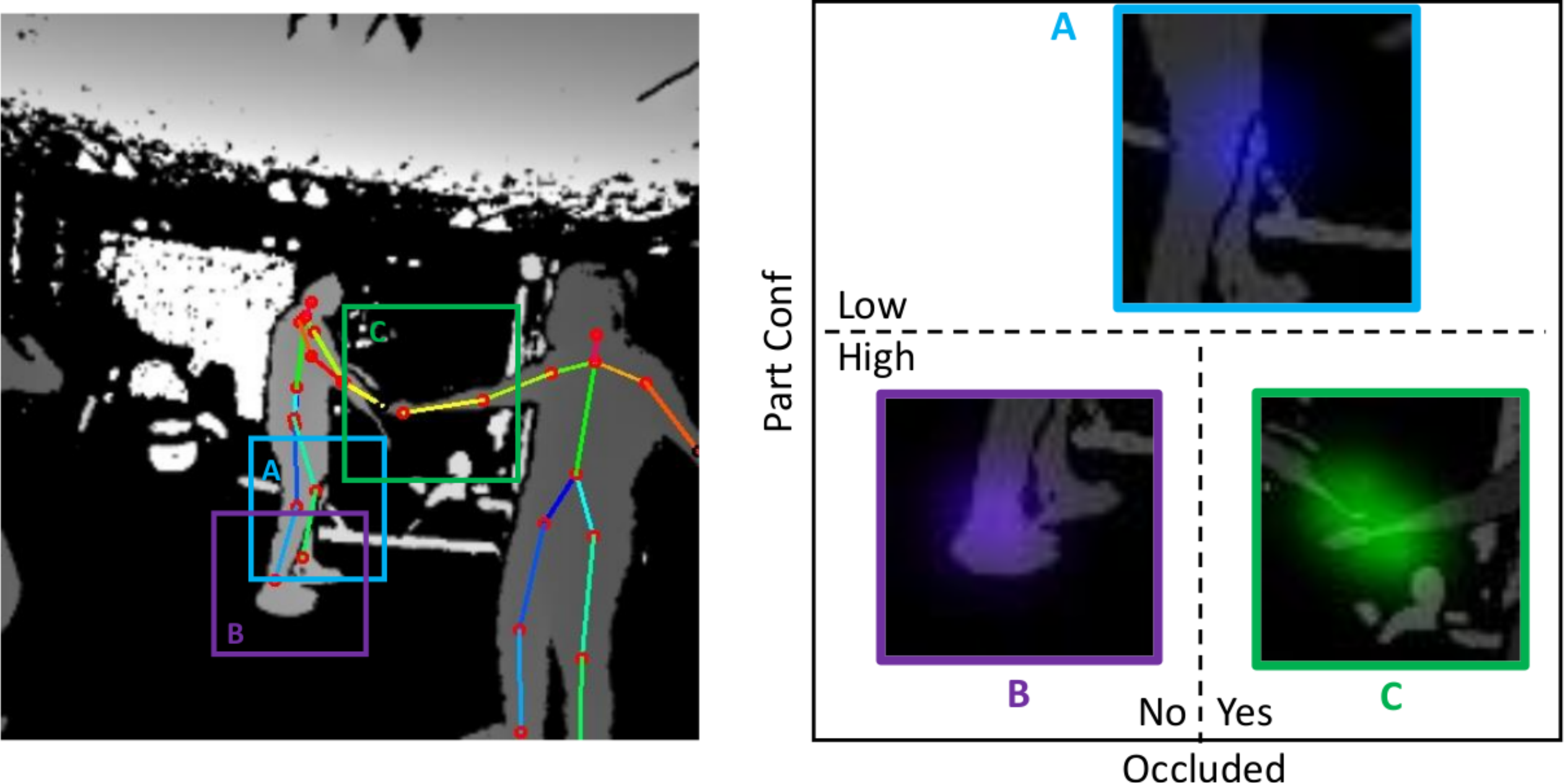}
    \caption{ \textbf{Conflicting cases to resolve in fusion.} Part confidence maps for the marked regions are visualized to illustrate three conflicting cases to resolve. A: The confidence of left knee is low. B: The confidence of right foot is high without ambiguity. C: The confidence of occluded right hand is high but hallucinated by the same part from another person.}
  \label{fig:conflict:cases}
\end{figure}

The architecture of each network component and the preparation process of ground-truth maps and weight maps are illustrated in detail in Appendix~\ref{sec:networks}. In summary, the backbone and the heatmap branch follow the structure proposed in the simplified OpenPose for depth data~\cite{Martiez-Gonzalez:etal:iros2018:rpm}. The depth and TPDF branches adopt the same number of layers as the heatmap branch but with customized feature dimensions to balance the efficiency and robustness. The global pose network follows a similar architecture as the layers after the backbone in Yolo2~\cite{Redmon:Farhadi:cvpr2017:yolo2}, and uses a similar anchor-based representation.

\subsection{Fusion process}

TPDF enables an explicit fusion of part representations and global poses. As illustrated in Figure~\ref{fig:main:idea}, a 2D part predicted from a global pose located at $(x_j, y_j)$ is updated to a new position $(\bar{x}_j, \bar{y}_j)$ following the displacement vector in the predicted TPDF of part $j$, such that $\bar{x}_j = x_j + X_j(x_j, y_j), \bar{y}_j = y_j + Y_j(x_j, y_j)$.
%\begin{align}
%    \bar{x}_j &= x_j + X_j(x_j, y_j)\\
%    \bar{y}_j &= y_j + Y_j(x_j, y_j).
%\end{align}
To improve accuracy, weighted aggregation is later applied to estimate the final 2D position $\{(\hat{x}_j, \hat{y}_j)\}^K_{j=1}$ and depth $\{\hat{Z_j}\}^K_{j=1}$, as illustrated in Figure~\ref{fig:main:idea}. Specifically, $X_j$, $Y_j$, and $D_j$ within a mask $M$ centered at the updated integer position $(\lfloor\bar{x}_j\rfloor, \lfloor\bar{y}_j\rfloor)$ is averaged by using $H_j$ as aggregation weights, which leads to the following equations:
\begin{align}
    \hat{x}_j &= \lfloor\bar{x}_j\rfloor + \frac{\sum_{(u, v)\in M} H_j(u, v) \cdot X_j(u, v)}{\sum_{(u, v)\in M} H_j(u, v)}\\
    \hat{y}_j &= \lfloor\bar{y}_j\rfloor + \frac{\sum_{(u, v)\in M} H_j(u, v) \cdot Y_j(u, v)}{\sum_{(u, v)\in M} H_j(u, v)}\\
    \hat{Z}_j &= \frac{\sum_{(u, v)\in M} H_j(u, v) \cdot D_j(u, v)}{\sum_{(u, v)\in M} H_j(u, v)}.
\end{align}
Predicted $\{(\hat{x}_j, \hat{y}_j)\}^K_{j=1}$ and $\{\hat{Z_j}\}^K_{j=1}$ are transformed to 3D positions given known camera intrinsic parameters.

\subsection{Resolving conflicting cases}

%Due to different occlusion cases, global poses may appear to be in conflict with part predictions, therefore determining the reliable source of prediction is another key issue to resolve in the fusion process. 
%However, the presented fusion process only works when a part from global poses falls within the truncated range of the corresponding TPDF. Meanwhile, 
There are conflicting cases when multiple human bodies occlude each other or a global pose falls out of the effective range of a TPDF. To resolve them, a mode selection scheme is carefully designed. The scheme utilizes the part confidence maps $\{H_j\}^{K+1}_{j=1}$ from the heatmap branch and the part visibility attributes from the global pose network.

\begin{figure*}[!h]
    \centering
    \includegraphics[width=0.95\textwidth]{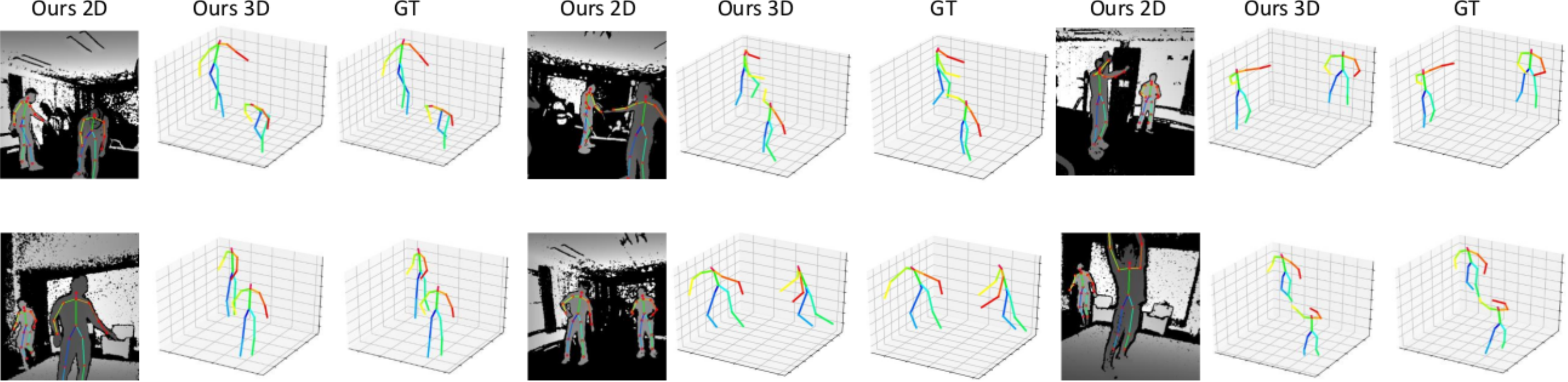}
    \caption{ \textbf{PoP-Net visual results.} Predictions in 2D, 3D, and the ground-truth are visualized on six examples from MP-3DHP.}
  \label{fig:MP-3DHP:popnet:vis}
\end{figure*}

As illustrated in Figure~\ref{fig:conflict:cases}, there are in total three cases to consider respectively: (A) when the part confidence $H_j$ is low at a global part position, the global detection is used directly, which is usually observed when the position of a part is not accessible due to truncation or occlusion; (B) when the part confidence is high and no occlusion from another instance of the same part is involved, the presented fusion process is applied; and (C) a challenging case may occur when the part confidence is high but is impacted by occlusion from another instance of the same part type. Fortunately, since the part depth map is prepared following a \textit{z}-buffer rule, a significant difference between the global part depth and the part depth map will be observed in this case. Therefore, part visibility attributes $\{v_j\}^K_{j=1}$ can be used as indicators of case (C), which can be integrated into the global pose representation. At last, the visual results of PoP-Net are shown in Figure~\ref{fig:MP-3DHP:popnet:vis} on a set of multi-person testing samples. %this case can be detected by introducing additional part visibility $\{v_j\}^K_{j=1}$ to the global pose representation.

%%%%%%%%%%%%%%%%%%%%%%%%%%%%%%%%% A better design may worth a try %%%%%%%%%%%%%%%%%%%%%%%%%%%%%%%%%%%%%%%%%%%%%%%%%%%%

% (\textcolor{red}{A better way not implemented yet: However, separate supervision for different branches may cause positional incompatibility in applying aggregation. To solve this issue, we design a new loss to enforce the aggregation weight map within the foreground region to align with depth encoding and TPDF encoding, and in an explicit way. Specifically, similar to the heatmap loss, the ground-truth value within background region is set to be zero. However, the foreground values are not explicitly trained through the distance of heatmap, but are learned via as L2 regularization. As shown in Equation..., the weights learned from L2 regularization indeed measures the uncertainty of local depth encoding and TPDF encoding. An unique benefit from this design compared to direct heatmap implementation lies on the capability in reasoning occluded part. The network could learn to adjust the value from the context to approximate the right value associated with an occluded body part.})

%%%%%%%%%%%%%%%%%%%%%%%%%%%%%%%%%%%%%%%%%%%%%%%%%%%%%%%%%%%%%%%%%%%%%%%%%%%%%%%%%%%%%%%%%%%%%%%%%%%%%%%%%%%%%%%%%%%%%%

%%%%%%%%%%%%%%%%%%%%%%%%%%%%%%%%%%%%%%%%%%%%%%%%%%%%%%%%%%%%%%%%%%%%%%%%%%%%%%%%%%%%%%%%%%

\section{MP-3DHP: Multi-Person 3D Human Pose Dataset}

Due to the lack of high-quality depth datasets for 3D pose estimation, we constructed Multi-Person 3D Human Pose Dataset (MP-3DHP) to facilitate the development of 3D pose estimation targeting real-world multi-person challenges. There are a few existing depth datasets for human pose estimation, but the data quality and diversity is rather limited. DIH~\cite{Martiez-Gonzalez:etal:iros2018:rpm} and K2HPD~\cite{Wang:etal:ACM2016} include a decent amount of data but are limited to 2D poses. ITOP~\cite{Haque:etal:eccv2016:itop} is a widely tested depth dataset for 3D pose estimation. However, data from ITOP is strictly limited to single person, clean background, and low diversity in object scales, camera angles, and pose types.

\begin{figure*}[!h]
    \centering
    \includegraphics[width=0.16\textwidth]{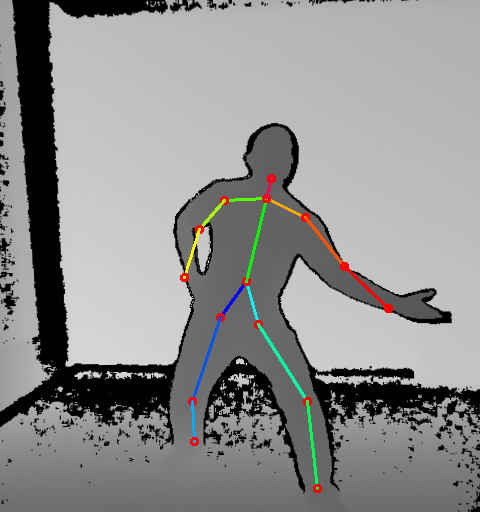}
    \includegraphics[width=0.16\textwidth]{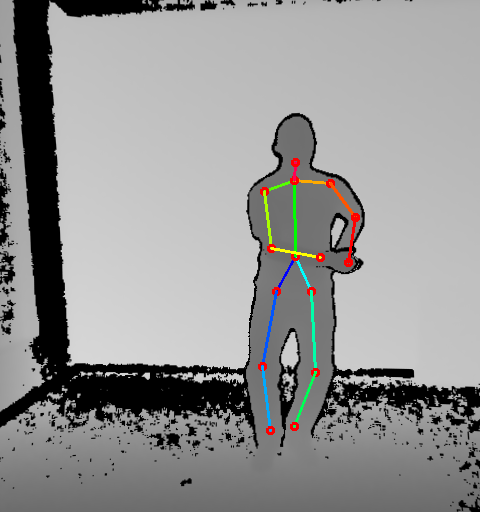}
    \includegraphics[width=0.16\textwidth]{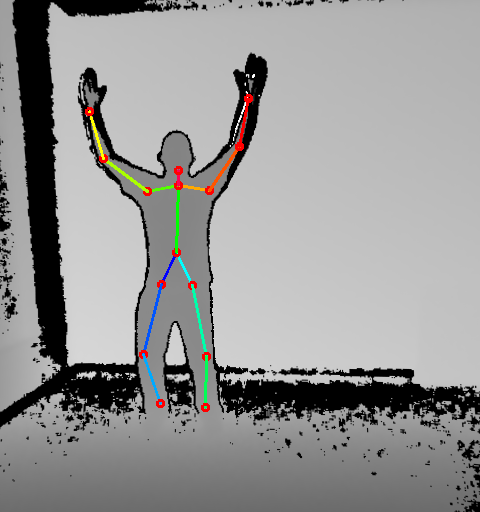}
    \includegraphics[width=0.16\textwidth]{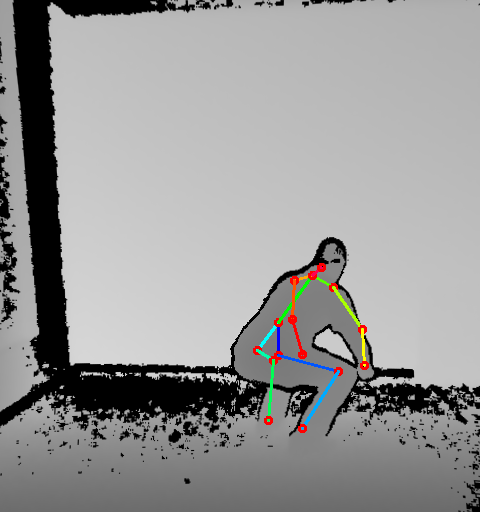}
    \includegraphics[width=0.16\textwidth]{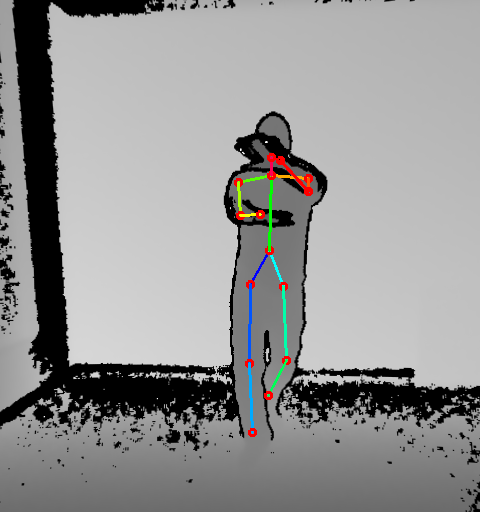}\\
    \includegraphics[width=0.16\textwidth]{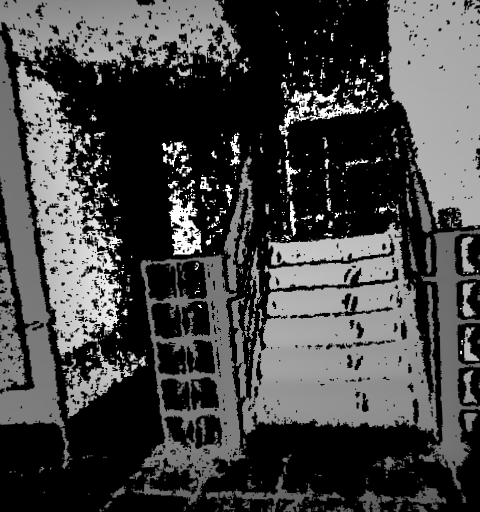}
    \includegraphics[width=0.16\textwidth]{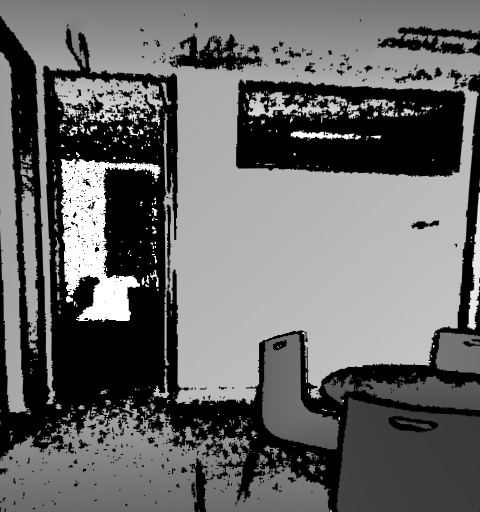}
    \includegraphics[width=0.16\textwidth]{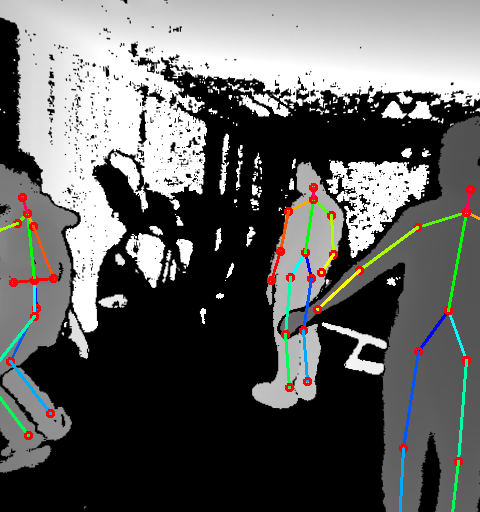}
    \includegraphics[width=0.16\textwidth]{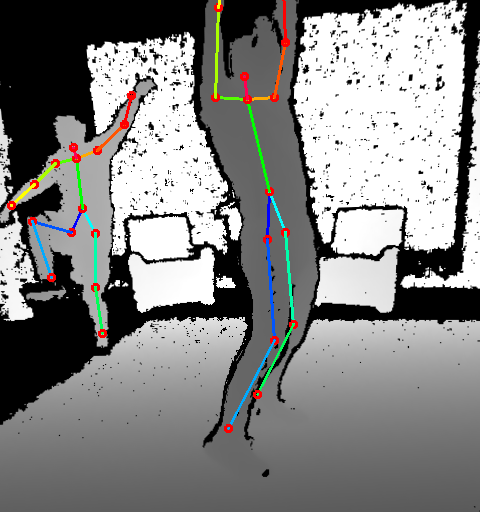}
    \includegraphics[width=0.16\textwidth]{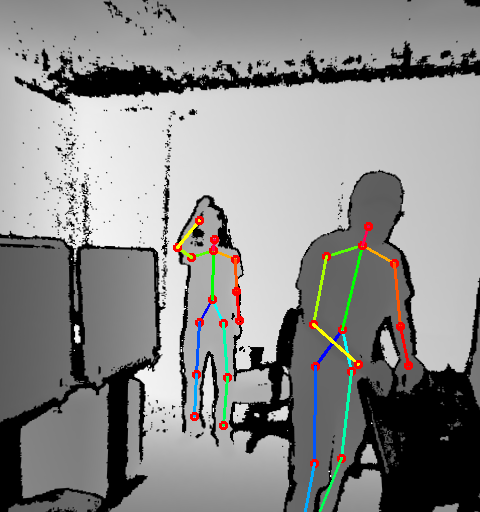}
    \caption{\textbf{ MP-3DHP examples.} (Top) Five single-person examples recorded from different locations from the training set. (Bottom) Two examples of background scenes, and three examples from multi-person testing set.}
  \label{fig:MP-3DHP:dataset}
\end{figure*}

MP-3DHP is designed to effectively cover the essential variations in human poses, object scales, camera angles, truncation scenarios, background scenes, and dynamic occlusion. Because collecting sufficient data to fully represent multi-person configurations combined with different background scenes is intractable due to combinatorial explosion, real data is only collected to cover the variations not achievable from data composition or data augmentation. Specifically, our training data collection focuses on single-person data involving varying poses, different object scales, varying camera ray angles, and additional background-only data covering different types of scenes. The remaining types of data variation are covered via data composition and data augmentation utilizing collected human segments. The testing set focuses on multi-person data under uncontrolled real-world scenarios. Figure~\ref{fig:MP-3DHP:dataset} shows examples from the training set, the background scenes and the multi-person testing set respectively.

\begin{comment}
\begin{table*}[]
\centering
\begin{tabular}{|l|l|l|l|l|l|l|l|l|l|}
\hline
\textbf{Dats Set} & \textbf{Images} & \textbf{People} & \textbf{\# CL} & \textbf{\# CR} & \textbf{\# FL} & \textbf{\# FR} & \textbf{\# Free} & \textbf{Seg} & \textbf{MP} \\ \hline
train             & 176828          & 13              & 36427          & 35921          & 38543          & 53728          & 12209            & yes          & no          \\ \hline
val               & 32719           & 2               & 6485           & 7093           & 5975           & 9778           & 3388             & yes          & no          \\ \hline
bg                & 8680            & 0               & 0              & 0              & 0              & 0              & 0                & no           & no          \\ \hline
test              & 4484            & 5               & 0              & 0              & 0              & 0              & 4484             & no           & yes         \\ \hline
\end{tabular}
\caption{\textbf{Summary of MP-3DHP}}
\label{tab:dataset}
\end{table*}
\end{comment}

\begin{table}[h]
\centering
\begin{tabular}{|l|l|l|l|l|l|l|l|l|}
\hline
\textbf{Set} & \textbf{Img} & \textbf{Sbj} & \textbf{Loc} & \textbf{Ori} & \textbf{Act} & \textbf{Sn} & \textbf{L+} \\ \hline
train        & 176828       & 13           & 4+           & 4            & 10+          & 1           & seg          \\ \hline
val          & 32719        & 2            & 4+           & 4            & 10+          & 1           & seg          \\ \hline
bg           & 8680         & 0            & 0            & 0            & 0            & 8           & no           \\ \hline
test         & 4484         & 5            & 0            & 0            & free         & 4           & mp           \\ \hline
\end{tabular}
\caption{\textbf{MP-3DHP summary.} The total number of images (Img), human subjects (Sbj), recording locations (Loc), self-orientations (Ori), action types (Act), scenes (Sn) are summarized. Additional label type (L+) indicates whether a set has segmentation (seg) or multi-person (mp) labels.}
\label{tab:dataset}
\end{table}

\subsection{Construction procedure} 

We utilized Azure Kinect to record human depth videos and automatically extracted 3D human poses associated with each depth image. Overall, 20 human candidates were involved in the recording procedure; 15 of them were recorded individually to construct the training set, while the remaining five were recorded in multi-person sessions to produce the multi-person testing set. For the training set, each candidate was recorded with a clean background in four trials at four different locations within the camera frustum. In each trial, a candidate was asked to perform 10 predetermined actions while facing four different orientations spanning $360^\circ$ and an additional short sequence of free-style movements towards the end. A classic graph-cut based method was applied to produce human segments for the training set. In addition, background images were recorded separately with moderate camera movements from eight different scenes. For the testing set, the remaining five people were recorded while performing random actions with different combinations in four different scenes. Table~\ref{tab:dataset} shows the statistics of our MP-3DHP dataset.

To collect reliable 3D pose ground-truth, human annotations are integrated into our pseudo-automatic data collecting system to sift out those unqualified samples. Specifically, we used two calibrated and synchronized cameras mounted on a solid bar in data capture, one is from Azure Kinect and the other is from a commercial cellphone. A 3D pose output from Azure Kinect is only selected as ground-truth when its projections to both views are visually correct.

%%%%%%%%%%%%%%%%%%%%%%%%%%%%%%%%%%%%%%%%%%%%%%%%%%%%%%%%%%%%%%%%%%%%%%%%%%%%%%%%%%%%%%%

\section{Experiments}

We experimentally substantiate the superiority of PoP-Net in depth image-based multi-person 3D pose estimation on two datasets, under two evaluation metrics, compared to prior state-of-the-art methods.

\subsection{Experimental setup}

\subsubsection{Datasets}
A depth-based 3D pose estimation method is evaluated on two benchmarks: the MP-3DHP and ITOP~\cite{Haque:etal:eccv2016:itop} datasets. MP-3DHP includes highly diverse 3D human data and provides reliable human segments to enable background augmentation and multi-person augmentation. The evaluation on MP-3DHP aims to determine a method's capability in handling real-world challenges in multi-person 3D pose estimation. Meanwhile, ITOP is a widely tested dataset for single-person 3D pose under highly controlled environment. We report results on ITOP to compare with prior state-of-the-arts on a simplified task.

\subsubsection{Evaluation metric}

%Evaluation of multi-person methods not only needs to consider the quality of part precision but also the detection accuracy. 
A method is evaluated in PCK and mAP with different focuses. First, PCK is a measurement that focuses on pose estimation without considering redundant detections. In our experiments, PCK is calculated as the average percentage of accurate key points on the best-matched predictions to the ground-truth poses, where the best match is based on the IOU between 2D bounding boxes. Second, the mAP metric introduced in MPII~\cite{DBLP:conf/cvpr/AndrilukaPGS14:mpii} dataset is applied, which integrates both the object detection and pose estimation accuracy in an overall score. For both PCK and mAP, $0.5$-head size rule is applied for 2D while 10-cm rule is applied for 3D.

\subsubsection{Competing methods}
A few prototypical methods have been compared with our method: (a) Yolo-Pose+ represents a typical top-down method, which is a pose estimation network extended from yolo-v2~\cite{Redmon:Farhadi:cvpr2017:yolo2} and implemented by us; (b) Open-Pose+ is a pure bottom-up method that is an extension from RPM~\cite{Martiez-Gonzalez:etal:iros2018:rpm}, a simplified version of OpenPose~\cite{Cao:etal:cvpr17:openpose}) that we extended with a depth branch to predict 3D poses; (c) A2J~\cite{Xiong:etal:iccv2019:a2j} represents the state-of-the-art two-stage pose estimation from a depth image.

To conduct a fair comparison, Yolo-Pose+, Open-Pose+, and PoP-Net are implemented using as many identical modules as possible. Specifically, Yolo-Pose+ is composed of the backbone network, the global pose network used for PoP-Net, and five additional intermediate $3\times3$ convolutional layers with $256d$ features in between. Open-Pose+ integrates the same backbone network, heatmap branch, depth branch as PoP-Net, and an additional Part Affinity Field (PAF) branch proposed in~\cite{Cao:etal:cvpr17:openpose}. The post process of Open-Pose+ follows the original work to use bipartite matching upon PAF to assemble detected parts into human bodies, and in addition reads the depth branch output to produce 3D poses. For A2J~\cite{Xiong:etal:iccv2019:a2j}, we use the identical network presented in the original paper to reproduce the results. Because A2J needs to work with given bounding-boxes for the multi-person case, we provide the predicted bounding boxes from Yolo-Pose+ to A2J so that the bounding-box quality is comparable to the other methods.

\subsubsection{Implementation details}

The input depth images are resized to $224 \times 224$ for Yolo-Pose+, Open-Pose+, and PoP-Net. The images are cropped and resized to $288 \times 288$ for A2J. Yolo-Pose+, Open-Pose+, PoP-Net are trained via standard SGD optimizer for 100 epochs, while A2J is trained via Adam optimizer following the original paper. Yolo-Pose+, Open-Pose+, and PoP-Net use two anchors with size $6\times12$ and $3\times6$, respectively. Following the study provided in~\cite{Cao:etal:cvpr17:openpose}, all the functional branches both in PoP-Net and Open-Pose+ use $S=2$ stages for an optimal balance in efficiency and accuracy. PoP-Net uses the TPDF truncated range $r=2$ in training, an optimal setting found in ablation study. At inference, the aggregation mask $M$ used in fusion is practically defined as a $5 \times 5$ square. An identical data augmentation process is applied to the training of each method. The basic data augmentation process is applied to both MP-3DHP and ITOP, which includes random rotation, flipping, cropping, and a specific depth augmentation described in Appendix~\ref{sec:depth:aug}. While the multi-person augmentation is only applied on MP-3DHP. 

\subsection{MP-3DHP dataset}

On MP-3DHP dataset, each method is trained on the training set with multi-person augmentation on top of basic data augmentation. Given the provided human segments, background augmentation can be applied by superimposing the human mask region from a training image onto a randomly selected background image. Meanwhile, multi-person augmentation can also been applied by superimposing multiple human segments onto a random background scene following \textit{z}-buffer rule, which is described in more detail in Appendix~\ref{sec:mpaug}. %The background augmentation aims to improve a learned model's robustness to new background. The multi-person augmentation in addition aims to improve the robustness under interactive occlusion between multiple human instances. 
In testing, a method is evaluated on four different datasets representing different levels of challenges: (1) the validation set (\textbf{Simple}), (2) the background augmented (\textbf{BG Aug}) set constructed from validation set , (3) the multi-person augmented (\textbf{MP Aug}) set constructed from validation set, and (4) the real multi-person (\textbf{MP Real}) testing set including challenging real-world  recordings.

\subsubsection{Quantitative comparison} 

PoP-Net is compared with the other methods on MP-3DHP. As observed in Table~\ref{tab:MP-3DHP:eval}, PoP-Net achieves the state-of-the-art almost on every testing set, and significantly surpasses other methods under the most challenging metric: 3D mAP on the MP Real test. Open-Pose+ shows marginal advantages in 2D mAP in certain cases because it could benefit from higher recall by recovering isolated parts without seeing a whole body. However Open-Pose+ is more erroneous in depth prediction under occlusion due to its pure bottom-up pipeline, hence its 3D mAP drops significantly. A2J, on the other hand, shows marginal advantage in 3D PCK in certain tests, which can be interpreted as that the global weighted aggregation could leverage the full context within ROI to infer the depth of an occluded part. However, A2J appears to be rather sensitive to the quality of predicted ROIs; therefore its mAP performance is not optimal. %A2J also requires to process each ROI individually, thus its computational cost scales up with the number of human bodies appeared in an image.

%Visual results of PoP-Net are shown in Figure~\ref{fig:MP-3DHP:popnet:vis} for the real testing set, while additional qualitative comparisons on the most challenging cases are provided in Appendix~\ref{sec:vis:hard}.

\begin{table}[h]
\centering
\setlength{\tabcolsep}{2pt}
\renewcommand{\arraystretch}{1.0}
\begin{tabular}{|l|l|l|l|l|l|}
\hline
Test                     & Method          & 2D PCK          & 3D PCK          & 2D mAP          & 3D mAP          \\ \hline
\multirow{4}{*}{Simple}  & Yolo-Pose+      & 0.957           & 0.910           & 0.926           & 0.847           \\ \cline{2-6} 
                         & Open-Pose+      & \blue{0.967}    & 0.915           & \blue{0.967}    & \blue{0.893}    \\ \cline{2-6} 
                         & Yolo-A2J        & 0.959           & \blue{0.924}    & 0.936           & 0.868           \\ \cline{2-6} 
                         & PoP-Net         & \textbf{0.978}  & \textbf{0.947}  & \textbf{0.974}  & \textbf{0.926}  \\ \hline
\multirow{4}{*}{BG Aug}  & Yolo-Pose+      & 0.956           & 0.904           & 0.923           & 0.834           \\ \cline{2-6} 
                         & Open-Pose+      & 0.911           & 0.916           & \blue{0.969}    & \blue{0.885}    \\ \cline{2-6} 
                         & Yolo-A2J        & \blue{0.964}    & \blue{0.927}    & 0.941           & 0.871           \\ \cline{2-6} 
                         & PoP-Net         & \textbf{0.982}  & \textbf{0.947}  & \textbf{0.977}  & \textbf{0.924}  \\ \hline
\multirow{4}{*}{MP Aug}  & Yolo-Pose+      & 0.872           & 0.777           & 0.799           & 0.651           \\ \cline{2-6} 
                         & Open-Pose+      & \blue{0.887}    & 0.765           & \textbf{0.870}  & 0.667           \\ \cline{2-6} 
                         & Yolo-A2J        & 0.876           & \textbf{0.819}  & 0.803           & \blue{0.707}    \\ \cline{2-6} 
                         & PoP-Net         & \textbf{0.906}  & \blue{0.808}    & \blue{0.863}    & \textbf{0.708}  \\ \hline
\multirow{4}{*}{MP Real} & Yolo-Pose+      & 0.734           & 0.607           & 0.616           & 0.449           \\ \cline{2-6} 
                         & Open-Pose+      & 0.805           & 0.641           & \textbf{0.802}  & 0.558           \\ \cline{2-6} 
                         & Yolo-A2J        & \blue{0.837}    & \textbf{0.724}  & 0.744           & \blue{0.574}    \\ \cline{2-6} 
                         & PoP-Net         & \textbf{0.839}  & \blue{0.708}    & \blue{0.799}    & \textbf{0.606}  \\ \hline
\end{tabular}
\caption{\textbf{Evaluation on MP-3DHP.} Competing methods are evaluated on four testing sets. The best method is marked in bold black while the second best is marked in blue.}
\label{tab:MP-3DHP:eval}
\end{table}

\subsubsection{Qualitative comparison}
\label{sec:vis:hard}

\begin{figure*}[!h]
    \centering
    (1) \includegraphics[width=0.185\textwidth]{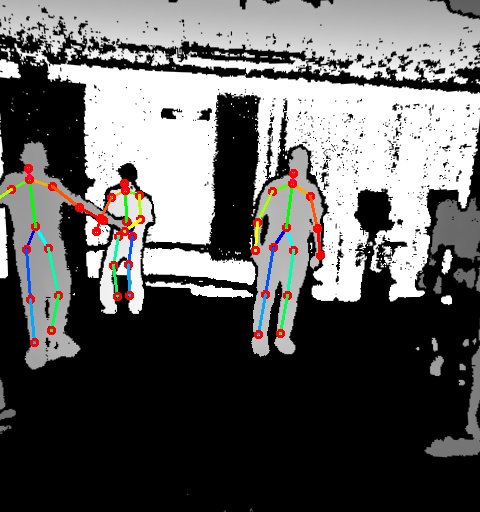}
    \includegraphics[width=0.185\textwidth]{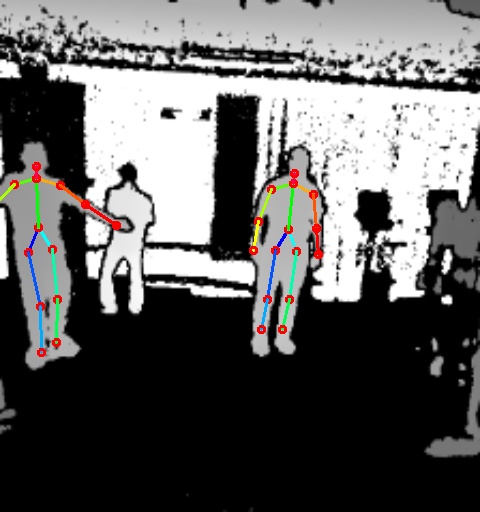}
    \includegraphics[width=0.185\textwidth]{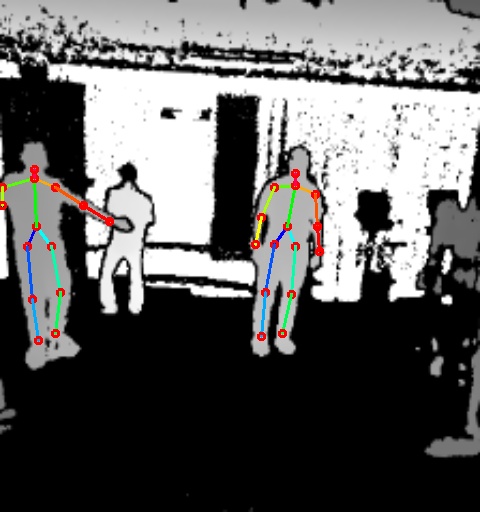}
    \includegraphics[width=0.185\textwidth]{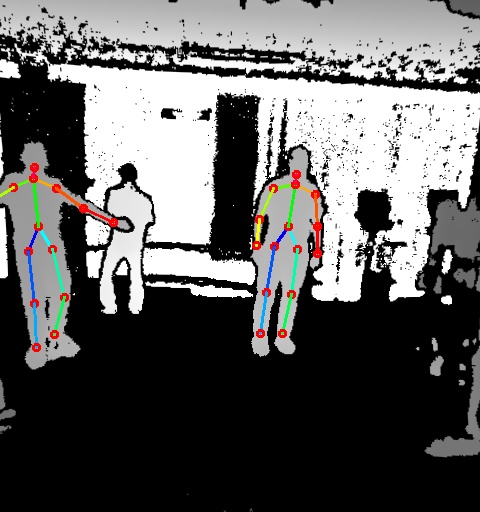}
    \includegraphics[width=0.185\textwidth]{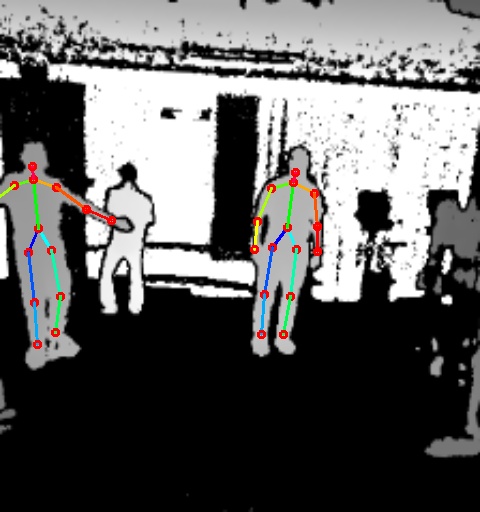} \\
    (2) \includegraphics[width=0.185\textwidth]{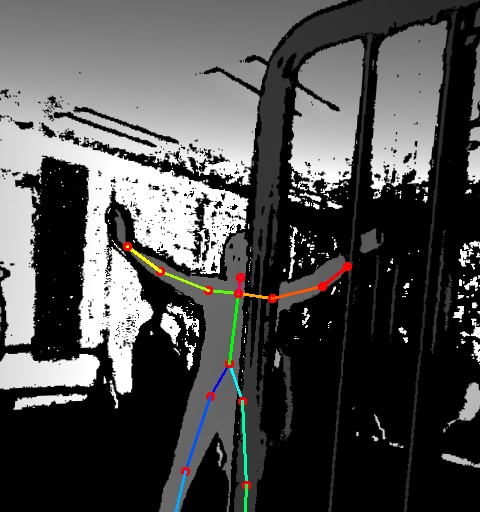}
    \includegraphics[width=0.185\textwidth]{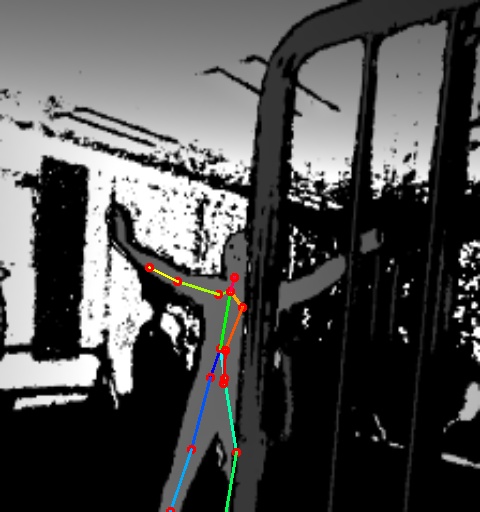}
    \includegraphics[width=0.185\textwidth]{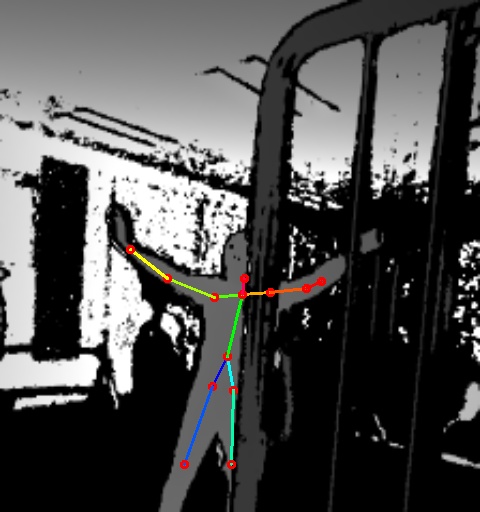}
    \includegraphics[width=0.185\textwidth]{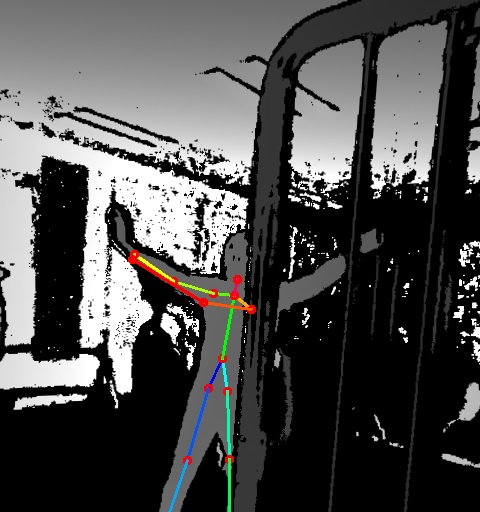}
    \includegraphics[width=0.185\textwidth]{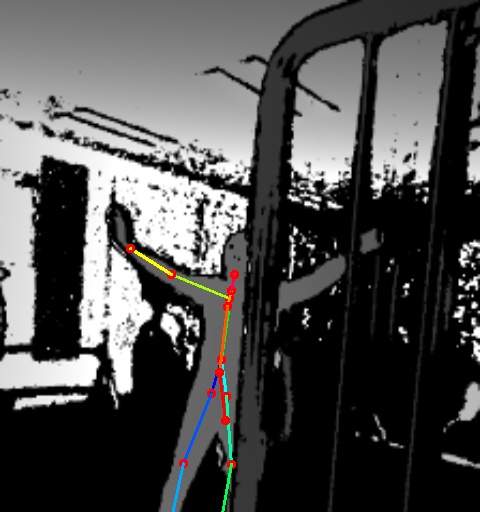} \\
    (3) \includegraphics[width=0.185\textwidth]{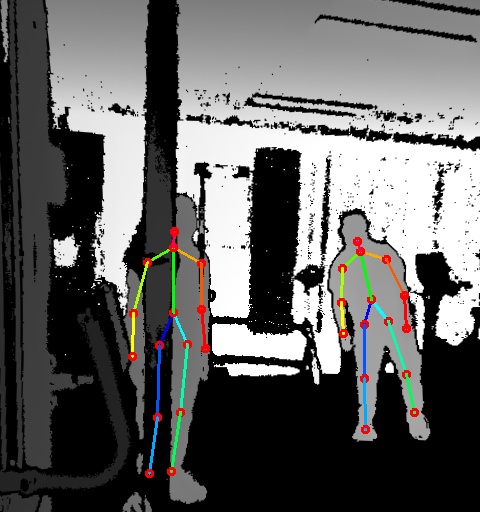}
    \includegraphics[width=0.185\textwidth]{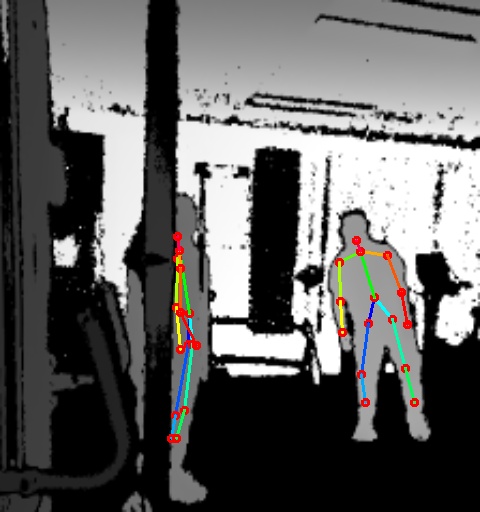}
    \includegraphics[width=0.185\textwidth]{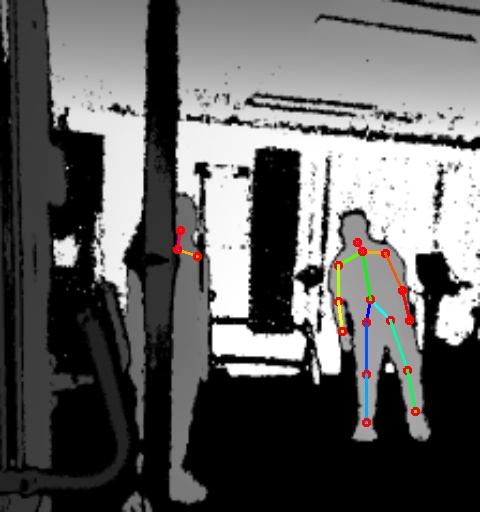}
    \includegraphics[width=0.185\textwidth]{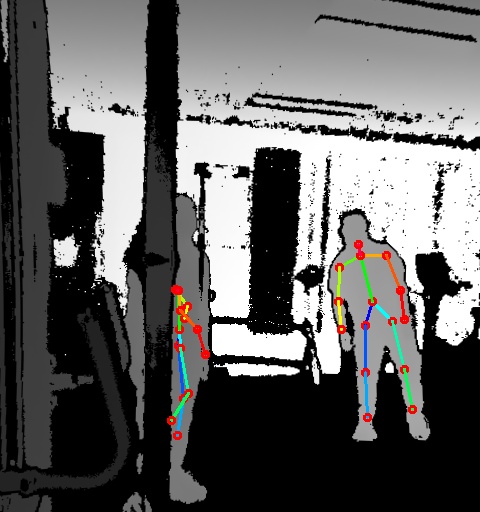}
    \includegraphics[width=0.185\textwidth]{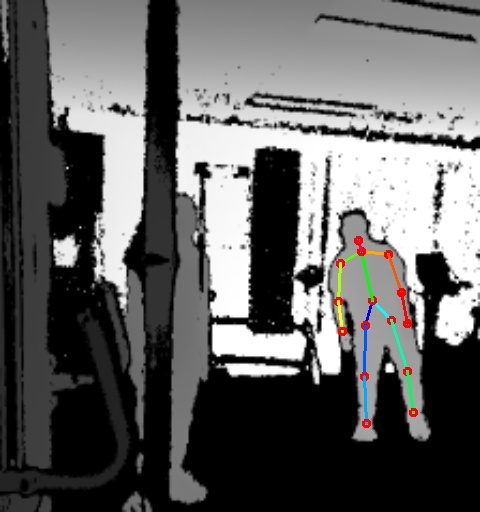} \\
    (4) \includegraphics[width=0.185\textwidth]{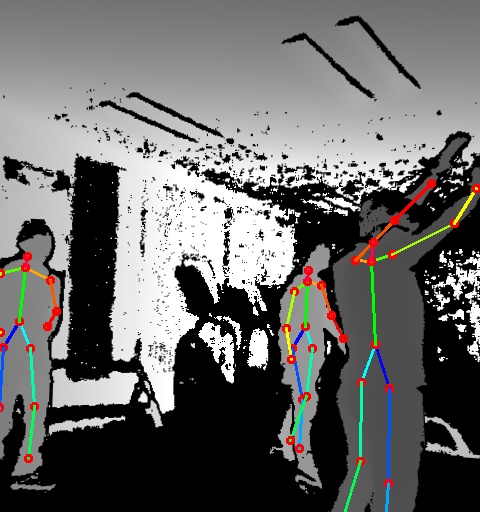}
    \includegraphics[width=0.185\textwidth]{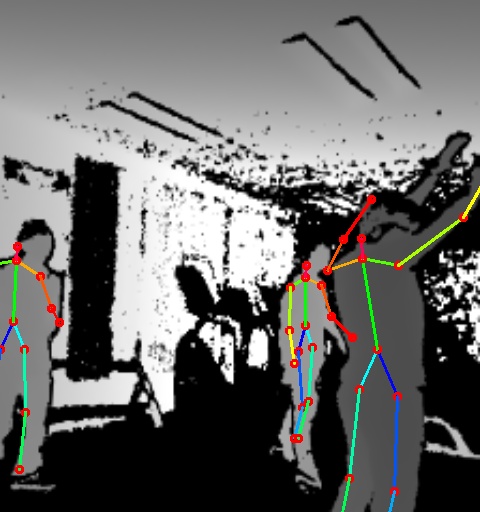}
    \includegraphics[width=0.185\textwidth]{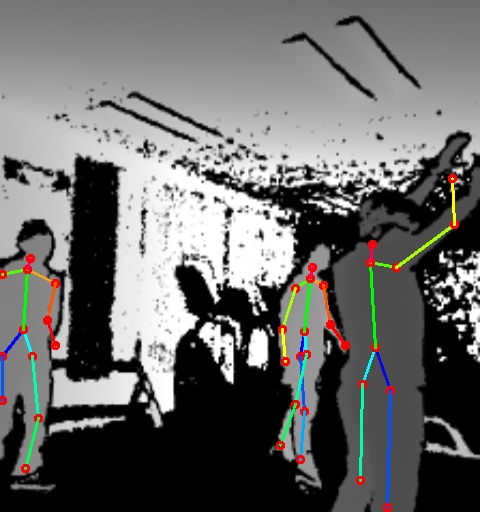}
    \includegraphics[width=0.185\textwidth]{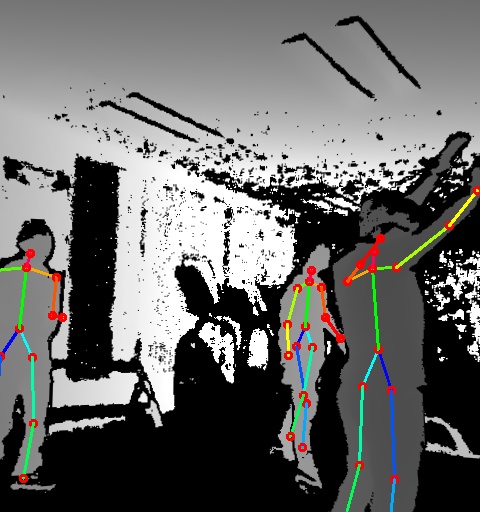}
    \includegraphics[width=0.185\textwidth]{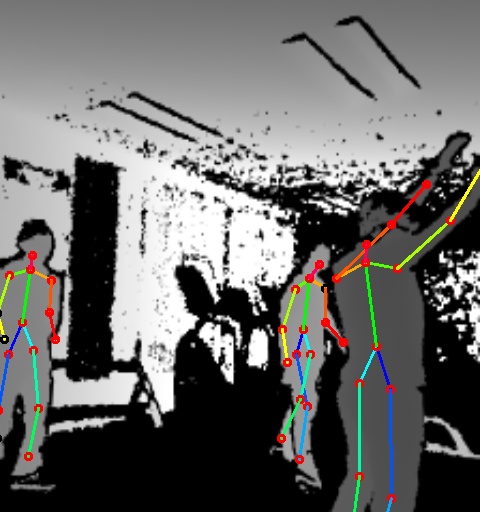} \\
    (5) \includegraphics[width=0.185\textwidth]{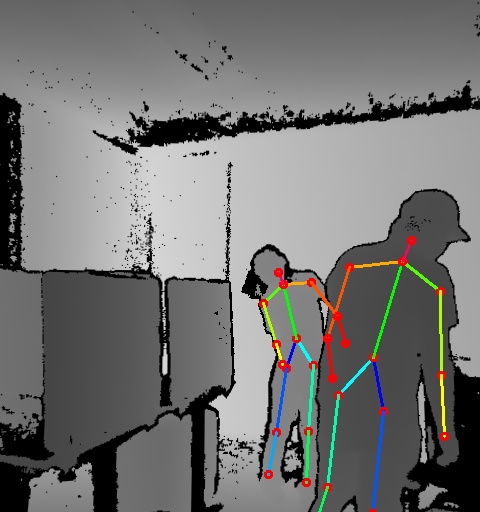}
    \includegraphics[width=0.185\textwidth]{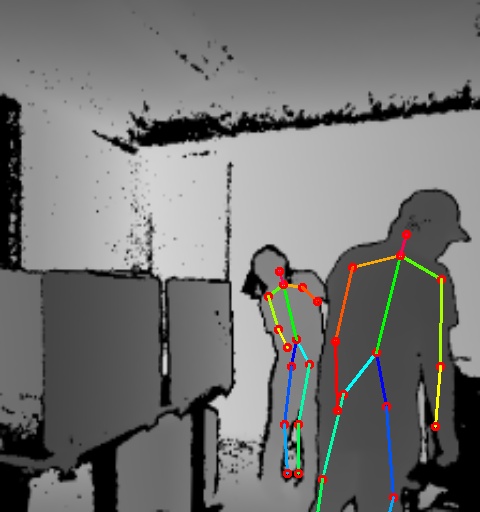}
    \includegraphics[width=0.185\textwidth]{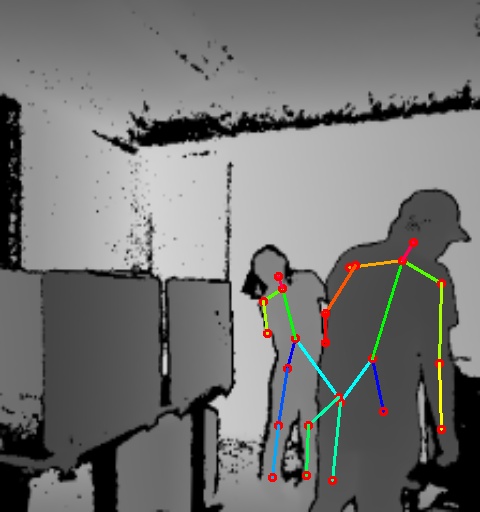}
    \includegraphics[width=0.185\textwidth]{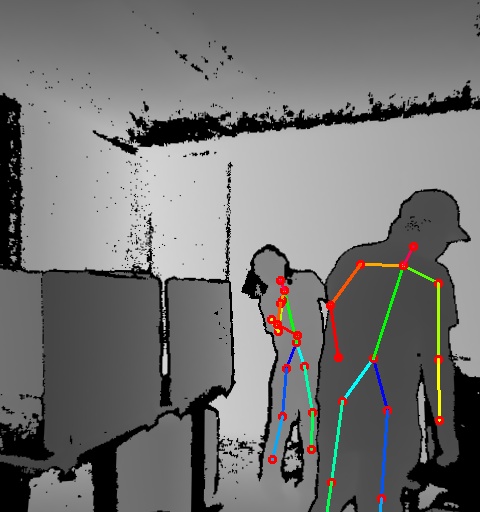}
    \includegraphics[width=0.185\textwidth]{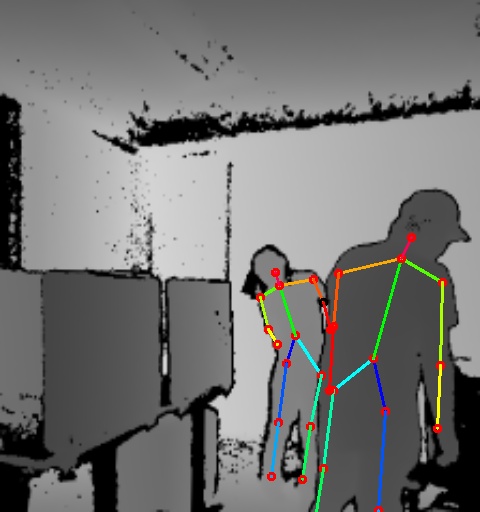} \\
    (6) \includegraphics[width=0.185\textwidth]{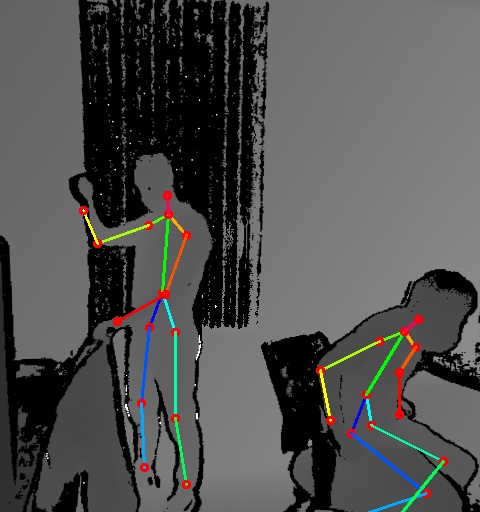}
    \includegraphics[width=0.185\textwidth]{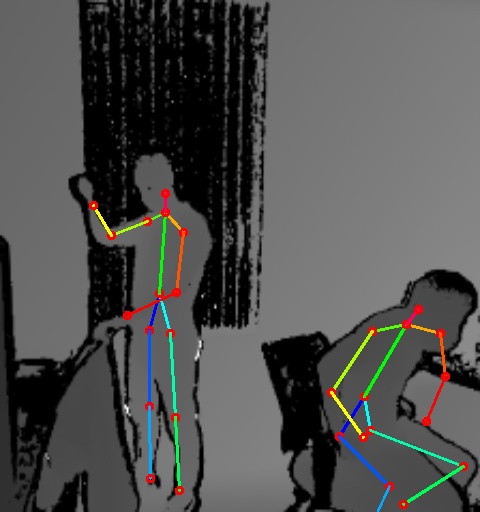}
    \includegraphics[width=0.185\textwidth]{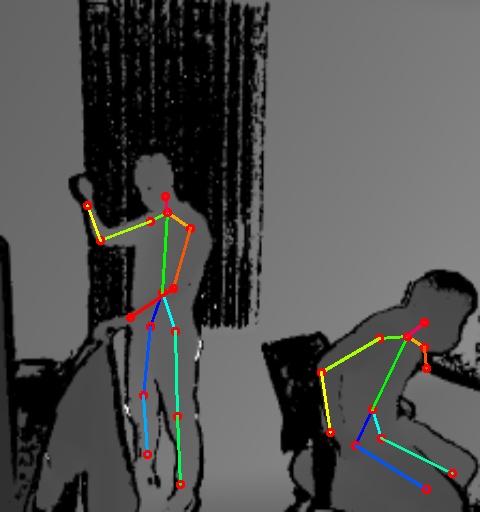}
    \includegraphics[width=0.185\textwidth]{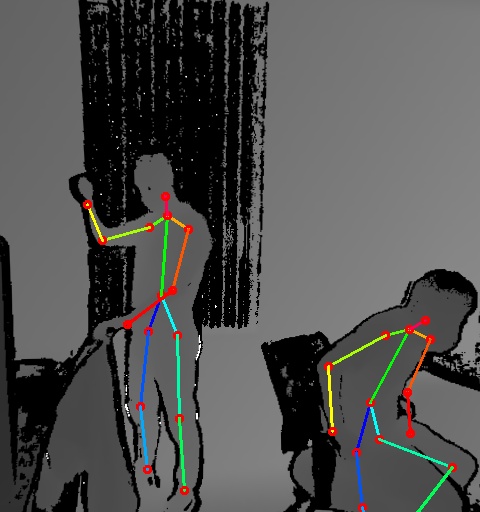}
    \includegraphics[width=0.185\textwidth]{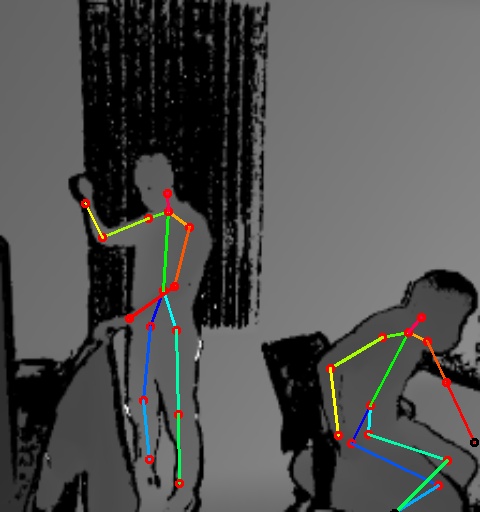} \\
     Ground-Truth \qquad\qquad\quad Yolo-Pose+ \qquad\qquad\qquad Open-Pose+ \qquad\qquad\qquad Yolo-A2J \qquad\qquad\qquad PoP-Net
\caption{\bf Visual comparison of competing methods on challenging cases.}
 \label{fig:visual:compare:hard}
\end{figure*}

In order to demonstrate that PoP-Net achieves the state-of-the-art and the proposed MP-3DHP represents real-world challenges, we compare the predicted poses from competing methods on a set of challenging cases. As shown in Figure~\ref{fig:visual:compare:hard}, we demonstrate visual comparison on: (1) an example including a target human captured far beyond the observation range in the training; (2-3) examples having severe background occlusion; (4-5) examples including multi-person occlusion and considerable truncation; and (6) an example with unobserved poses from training. As observed, PoP-Net in general is more reliable across all these cases. However, all methods failed on some most challenging cases. Such observation indicates that there is still huge room for improvement towards a robust approach in real-world challenges.

\subsection{Ablation study}
\label{sec:detail:ablation}

An ablation study has been conducted on MP-3DHP to analyze the effectiveness of PoP-Net components and the specific data augmentation methods. Because our motivation is to solve multi-person 3D pose estimation, we only conduct the ablation study on MP-3DHP, which to our knowledge is the only dataset providing multi-person 3D pose labels.

%\subsubsection{Ablation study}
\subsubsection{Effectiveness of fusing pose and parts}

The effectiveness of fusing part representations and global poses has been studied. As shown in Table~\ref{tab:pop:net:ablation}, evaluation has been done on four different testing sets from MP-3DHP, and separately on: (1) the 2D global poses predicted from the global pose network (\textbf{2D Glb}), (2) the final 2D poses after fusion (\textbf{2D Fuse}), (3) the 3D global poses predicted from the global pose network (\textbf{3D Glb}), and (4) the 3D poses computed from 2D fused poses and predicted depth of parts (\textbf{3D Fuse}). In addition, the upper bound of 3D poses computed from ground-truth 2D poses and predicated depth (\textbf{3D UB}) has been reported to illustrate the importance of depth prediction. 

As observed, 2D and 3D poses after fusion constantly improve upon the direct predictions from the global pose network, and the margin increases on more challenging testing sets. Meanwhile, the accuracy of 3D pose prediction drops more significantly compared with 2D pose prediction on more challenging sets, where the upper bound is still far from ideal. Improving depth prediction under multi-person occlusion should be a main focus in a future work.

\begin{table}[h]
\setlength{\tabcolsep}{2pt}
\centering
\begin{tabular}{|l|l|l|l|l|l|l|}
\hline
Metric               & Test    & 2D Glb & 2D Fuse & 3D Glb & 3D Fuse & 3D UB \\ \hline
\multirow{4}{*}{PCK} & Simple  & 0.968  & 0.978   & 0.939  & 0.947   & 0.956 \\ \cline{2-7} 
                     & BG Aug  & 0.970  & 0.982   & 0.938  & 0.947   & 0.953 \\ \cline{2-7} 
                     & MP Aug  & 0.898  & 0.906   & 0.794  & 0.808   & 0.846 \\ \cline{2-7} 
                     & MP Real & 0.798  & 0.839   & 0.681  & 0.708   & 0.756 \\ \hline
\multirow{4}{*}{mAP} & Simple  & 0.963  & 0.974   & 0.917  & 0.926   & 0.931 \\ \cline{2-7} 
                     & BG Aug  & 0.965  & 0.977   & 0.915  & 0.924   & 0.927 \\ \cline{2-7} 
                     & MP Aug  & 0.849  & 0.863   & 0.701  & 0.708   & 0.735 \\ \cline{2-7} 
                     & MP Real & 0.755  & 0.799   & 0.582  & 0.606   & 0.622 \\ \hline
\end{tabular}
\caption{{\bf Ablation study on fusing pose and parts.}}
\label{tab:pop:net:ablation}
\end{table}

% 2D global pose (2D Glb), 2D fused output (2D Fuse), 3D global pose (3D Glb), 3D fused output (3D Fuse) and 3D pose upper unbound (3D UB) are reported on four testing sets, in PCK and mAP, respectively.

\subsubsection{Effectiveness of data augmentation}

The effectiveness of the depth augmentation (\textbf{D Aug}) method and the composition augmentation (\textbf{C Aug}) are analyzed. The depth augmentation considers different ranges of $a$. The composition augmentation includes background augmentation (\textbf{BG Aug}) and multi-person augmentation (\textbf{MP Aug}). Experiments have been conducted with a focus on the multi-person real testing set as these augmentation methods were motivated towards this ultimate task. However, the data augmentation methods are not limited to PoP-Net, but applicable to any method trained on MP-3DHP.

\begin{table}[h]
\setlength{\tabcolsep}{2pt}
\centering
\begin{tabular}{|l|l|l|l|l|l|}
\hline
D Aug               & C Aug              & 2D PCK         & 3D PCK         & 2D mAP         & 3D mAP         \\ \hline
0.7-1.7             & w\textbackslash{}o & 0.411          & 0.246          & 0.374          & 0.158          \\ \hline
0.7-1.7             & BG Aug             & 0.769          & 0.634          & 0.748          & 0.550          \\ \hline
\textbf{0.7-1.7}    & \textbf{MP Aug}    & \textbf{0.839} & \textbf{0.708} & \textbf{0.799} & \textbf{0.606} \\ \hline
w\textbackslash{}o  & MP Aug             & 0.610          & 0.481          & 0.617          & 0.427          \\ \hline
0.5-2.5             & MP Aug             & 0.835          & 0.648          & 0.785          & 0.508          \\ \hline
\end{tabular}
\caption{ {\bf Ablation study on data augmentation.}}
\label{tab:effect:aug}
\end{table}

As observed from Table~\ref{tab:effect:aug}, background augmentation leads to a significant improvement (over 30\%) compared to the baseline without any composition augmentation (2nd row vs. 1st row). Multi-person augmentation (3rd row) leads to another leap (about 5\%). On the other hand, the specific depth augmentation also plays a critical role in improving the robustness (about 18\% increase in 3D mAP, 3rd row  vs. 4th row), especially for objects from unobserved scales. However, further extension of the depth augmentation range leads to a drawback in 3D mAP (~10\%, 5th row vs. 3rd row), which is reasonable because the depth augmentation method can not fully simulate the data far beyond the original captured distance.

\subsubsection{Effectiveness of PoP-Net components}

To find an optimal configuration of the pipeline, we analyzed the effects of the truncated range $r$ on TPDF, the effectiveness of depth prediction (\textbf{D Pred}) and the conflict resolving scheme (\textbf{Sol C}), as reported in Table~\ref{tab:effect:pipeline}. %All the ablation study have been conducted on the multi-person real testing set, and in PCK and mAP metrics, such that the methods are optimized towards the ultimate task.

\begin{table}[]
\setlength{\tabcolsep}{2pt}
\centering
\begin{tabular}{|l|l|l|l|l|l|l|}
\hline
Sol C              & D Pred             & TPDF           & 2D PCK         & 3D PCK         & 2D mAP         & 3D mAP         \\ \hline
w                  & w\textbackslash{}o & r = 2          & 0.839          & 0.418          & 0.799          & 0.293          \\ \hline
w\textbackslash{}o & w                  & r = 2          & 0.833          & 0.696          & 0.797          & 0.598          \\ \hline
w                  & w                  & r = 1          & 0.827          & 0.678          & 0.782          & 0.560          \\ \hline
\textbf{w}         & \textbf{w}         & \textbf{r = 2} & \textbf{0.839} & \textbf{0.708} & \textbf{0.799} & \textbf{0.606} \\ \hline
w                  & w                  & r = 3          & 0.825          & 0.681          & 0.777          & 0.567          \\ \hline
w                  & w                  & r = 5          & 0.821          & 0.670          & 0.780          & 0.557          \\ \hline
w                  & w                  & r = 10         & 0.797          & 0.654          & 0.755          & 0.543          \\ \hline
w                  & w                  & r = inf        & 0.647          & 0.549          & 0.581          & 0.443          \\ \hline
\end{tabular}
\caption{\bf Ablation study on PoP-Net components.}
\label{tab:effect:pipeline}
\end{table}

\begin{figure}[!h]
    \centering
    \includegraphics[width=0.16\textwidth]{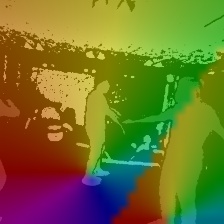}
    \includegraphics[width=0.16\textwidth]{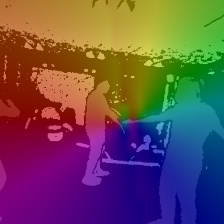}\\
    \includegraphics[width=0.16\textwidth]{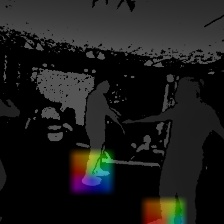}
    \includegraphics[width=0.16\textwidth]{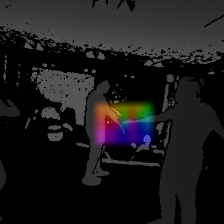}\\
    Right ankle \qquad\quad\qquad Right waist
  \caption{\textbf{ PDF vs. TPDF.} Full-range PDF examples (top) are compared with the TPDF (bottom) with truncated range $r=2$ at $\frac{h}{8} \times \frac{w}{8}$ feature resolution. Fields associated with the right ankle and right waist channels are visualized on the same depth image. }
  \label{fig:effect:trunction}
\end{figure}

A few important conclusions can be drawn from the results. First, depth prediction plays an important role in recovering reliable 3D poses, which leads to about 30\% improvements in 3D metrics compared with using the raw depth directly. Second, the introduced conflict resolving scheme leads to about 1\% improvement in 3D metrics. Third, applying an appropriate truncated range is critical in learning reliable models to predict part displacement vectors in multi-person scenarios. On the one hand, if the range is too limited, a global pose may not fall in the effective range of TPDF so that the positional accuracy can not be improved. On the other hand, a displacement field without truncation (r=inf) or with large truncated range would confuse the learning. As illustrated in Figure~\ref{fig:effect:trunction}, near those sharp boundaries, similar image patches could be associated with drastically different flow vectors. This leads to a degenerate regression problem.

\subsection{ITOP dataset}

PoP-Net is compared with competing methods on ITOP dataset. Because ITOP dataset is limited to single-person and clean background, PCK and mAP measurements are mostly identical. Therefore, we only report PCK metrics on ITOP dataset. To conduct a fair comparison, a method is trained and tested under two setups separately. One is based on provided ground-truth bounding boxes and the other directly uses the full image, as shown in Table~\ref{tab:itop:eval}. It can be observed that PoP-Net consistently outperforms Open-Pose+ and Yolo-Pose+ by a significant margin. Compared with A2J, PoP-Net is slightly worse in 3D and better in 2D, which is consistent with the PCK results reported on MP-3DHP. %Overall ITOP is not perfect for evaluating PoP-Net which is designed for multi-person.

\begin{table}[h]
\centering
\begin{tabular}{|l|l|l|l|}
\hline
Exp Setup                   & Method          & 2D PCK          & 3D PCK          \\ \hline
\multirow{2}{*}{GT Bbox}    & A2J             & 0.905           & \textbf{0.891}  \\ \cline{2-4} 
                            & PoP-Net         & \textbf{0.914}  & 0.882           \\ \hline
\multirow{4}{*}{Full Image} & Yolo-Pose+      & 0.833           & 0.787           \\ \cline{2-4} 
                            & Open-Pose+        & \blue{0.876}    & 0.778           \\ \cline{2-4} 
                            & Yolo-A2J             & 0.873           & \textbf{0.854}  \\ \cline{2-4} 
                            & PoP-Net         & \textbf{0.890}  & \blue{0.843}    \\ \hline
\end{tabular}
\caption{\textbf{Evaluation on ITOP dataset (front-view).} Methods are evaluated on ITOP dataset with and without GT bounding boxes.}
\label{tab:itop:eval}
\end{table}

\subsection {Running speed analysis}

\begin{comment}

The efficiency of each method is measured in FPS on multi-person test (2-3 people). The calculation of the running speed considers necessary post-processing to achieve the final set of multiple 3D poses and the bounding box prediction time for a two-stage method. As shown in Table~\ref{tab:efficiency}, the running speed of PoP-Net almost triples A2J and doubles Open-Pose+ on a single RTX 2080Ti GPU. The observation is as expected because OpenPose+ involves heavier post process and A2J's cost scales up with the number of humans. More detailed efficiency analysis is provided in Appendix~\ref{sec:detail:speed:analysis}.

\begin{table}[h]
\centering
\begin{tabular}{|l|l|l|l|l|}
\hline
           & Yolo-Pose+ & Open-Pose+ & A2J    & PoP-Net \\ \hline
FPS        & 223        & 48         & 32     & 91      \\ \hline
\end{tabular}
\caption{\bf Running speed on multi-person data.}
\label{tab:efficiency}
\end{table}

\section{Detailed Running Speed Analysis}
\label{sec:detail:speed:analysis}

\end{comment}

The efficiency of a method is measured in a few different metrics. First, the network complexity is measured in \textbf{MACs (G)}, which directly relates to the network inference time. Second, a method's average running time on an image including a single person (\textbf{SP}) is reported in milliseconds per image (\textit{ms}/\textit{im}). This metric considers not only network inference time, but also the essential pre-process to provide bounding boxes or the post-process to extract human poses. Third, a method's average running time on images including multiple people (\textbf{MP}) is also reported in milliseconds per image (\textit{ms}/\textit{im}). Finally, a method's average running speed on images including multiple people is measured by \textit{fps} which is equivalent to the metric in \textit{ms} per image on MP data. Every method has been tested on a single RTX 2080Ti GPU, and is evaluated in all the metrics as shown in Table~\ref{tab:run:time:detail}.

\begin{table}[h]
\setlength{\tabcolsep}{2pt}
\centering
\begin{tabular}{|l|l|l|l|l|}
\hline
                & Yolo-Pose+ & Open-Pose+ & A2J    & PoP-Net \\ \hline
MACs(G)         & 4.4        & 6.7        & 16.6   & 6.2     \\ \hline
SP (ms/im)      & 4.5        & 20         & 14     & 11      \\ \hline
MP (ms/im)      & 4.5        & 21         & 32     & 11      \\ \hline
MP (fps)        & 223        & 48         & 32     & 91      \\ \hline
\end{tabular}
\caption{\bf Running time analysis on multi-person data.}
\label{tab:run:time:detail}
\end{table}

As observed from MACs (G) scores, Yolo-Pose+ has the lightest network, while A2J has significantly more complex network compared with others. However, consider the pipeline running time, Open-Pose+ is much slower than the others on single-person images. This indicates that the part association post-process involved in Open-Pose+ is a much heavier process compared with the simple post-process used in Pop-Net. On the other hand, although A2J uses a more complex network, it almost has no post-processing cost so that its efficiency on single-person images is even better than Open-Pose+. Finally, as observed from multi-person pipeline running time and speed, the efficiency of A2J drops significantly while the other single-shot methods are not affected. Overall, PoP-Net shows significant advantages in efficiency, which almost triples A2J and doubles Open-Pose+ in multi-person scenarios.

%%%%%%%%%%%%%%%%%%%%%%%%%%%%%%%%%%%%%%%%%%%%%%%%%%%%%%%%%%%%%%%%%%%%%%%%%%%%%%%%%%%%%%%%%%%%%%%%%%%%%%%%%%%%%%%%%%%%%%%%%%%%%%
%%%%%%%%%%%%%%%%%%%%%%%%%%%%%%%%%%%%%%%%%%%%%%%%%%%%%%%%%%%%%%%%%%%%%%%%%%%%%%%%%%%%%%%%%%%%%%%%%%%%%%%%%%%%%%%%%%%%%%%%%%%%%%
%%%%%%%%%%%%%%%%%%%%%%%%%%%%%%%%%%%%%%%%%%%%%%%%%%%%%%%%%%%%%%%%%%%%%%%%%%%%%%%%%%%%%%%%%%%%%%%%%%%%%%%%%%%%%%%%%%%%%%%%%%%%%%
%%%%%%%%%%%%%%%%%%%%%%%%%%%%%%%%%%%%%%%%%%%%%%%%%%%%%%%%%%%%%%%%%%%%%%%%%%%%%%%%%%%%%%%%%%%%%%%%%%%%%%%%%%%%%%%%%%%%%%%%%%%%%%
%%%%%%%%%%%%%%%%%%%%%%%%%%%%%%%%%%%%%%%%%%%%%%%%%%%%%%%%%%%%%%%%%%%%%%%%%%%%%%%%%%%%%%%%%%%%%%%%%%%%%%%%%%%%%%%%%%%%%%%%%%%%%%

\section{Conclusion}

In this paper, we introduce PoP-Net for multi-person 3D pose estimation from a depth image. PoP-Net predicts part maps and global poses in a single pass and explicitly fuses them via utilizing the proposed Truncated Part Displacement Field (TPDF). Conflicting cases are effortlessly resolved in a rule-based process given part visibility and confidence out of the network. Meanwhile, a comprehensive 3D human depth dataset called MP-3DHP is constructed to facilitate the development of models for real-world multi-person challenges. In experiments, PoP-Net achieves state-of-the-art results on MP-3DHP and ITOP datasets with significant advantage in 3D mAP and running speed in processing multi-person data.\\

\noindent{\bf Acknowledgements}\\
This work was supported by OPPO US Research Center.

%\acknowledgments{
%This work was supported by OPPO US Research Center.}

%\cleardoublepage
\clearpage
%\newpage

%\bibliographystyle{abbrv}
\bibliographystyle{abbrv-doi}
%\bibliographystyle{abbrv-doi-narrow}
%\bibliographystyle{abbrv-doi-hyperref}
%\bibliographystyle{abbrv-doi-hyperref-narrow}

%\bibliography{newbib_short}
%\bibliography{newbib}

\begin{thebibliography}{10}

\bibitem{DBLP:conf/cvpr/AndrilukaPGS14:mpii}
M.~Andriluka, L.~Pishchulin, P.~V. Gehler, and B.~Schiele.
\newblock 2d human pose estimation: New benchmark and state of the art
  analysis.
\newblock In {\em {IEEE} Conference on Computer Vision and Pattern Recognition
  (CVPR)}, pp. 3686--3693, 2014.

\bibitem{BenzineCLPA:CVPR2020}
A.~Benzine, F.~Chabot, B.~Luvison, Q.~C. Pham, and C.~Achard.
\newblock Pandanet: Anchor-based single-shot multi-person 3d pose estimation.
\newblock In {\em 2020 {IEEE/CVF} Conference on Computer Vision and Pattern
  Recognition, {CVPR} 2020, Seattle, WA, USA, June 13-19, 2020}, pp.
  6855--6864. {IEEE}, 2020.

\bibitem{Cao:etal:cvpr17:openpose}
Z.~Cao, T.~Simon, S.~Wei, and Y.~Sheikh.
\newblock Realtime multi-person 2d pose estimation using part affinity fields.
\newblock In {\em IEEE Conference on Computer Vision and Pattern Recognition
  (CVPR)}, pp. 1302--1310, 2017.

\bibitem{Elhayek:etal:PAMI17}
A.~Elhayek, E.~de~Aguiar, A.~Jain, J.~Tompson, L.~Pishchulin, M.~Andriluka,
  C.~Bregler, B.~Schiele, and C.~Theobalt.
\newblock Marconi - convnet-based marker-less motion capture in outdoor and
  indoor scenes.
\newblock {\em {IEEE} Trans. Pattern Anal. Mach. Intell.}, 39(3):501--514,
  2017.

\bibitem{Haque:etal:eccv2016:itop}
A.~Haque, B.~Peng, Z.~Luo, A.~Alahi, S.~Yeung, and F.~Li.
\newblock Towards viewpoint invariant 3d human pose estimation.
\newblock In {\em 14th European Conference Computer Vision (ECCV)}, pp.
  160--177, 2016.

\bibitem{He:etal:iccv2017:mask:rcnn}
K.~He, G.~Gkioxari, P.~Doll{\'{a}}r, and R.~B. Girshick.
\newblock Mask {R-CNN}.
\newblock In {\em IEEE International Conference on Computer Vision (ICCV)}, pp.
  2980--2988, 2017.

\bibitem{He:etal:CVPR2016:resnet}
K.~He, X.~Zhang, S.~Ren, and J.~Sun.
\newblock Deep residual learning for image recognition.
\newblock In {\em IEEE Conference on Computer Vision and Pattern Recognition
  (CVPR)}, pp. 770--778, 2016.

\bibitem{DBLP:journals/pami/IonescuPOS14}
C.~Ionescu, D.~Papava, V.~Olaru, and C.~Sminchisescu.
\newblock Human3.6m: Large scale datasets and predictive methods for 3d human
  sensing in natural environments.
\newblock {\em {IEEE} Trans. Pattern Anal. Mach. Intell.}, 36(7):1325--1339,
  2014.

\bibitem{DBLP:conf/eccv/IqbalG16}
U.~Iqbal and J.~Gall.
\newblock Multi-person pose estimation with local joint-to-person associations.
\newblock In {\em European Conference on Computer Vision Workshops}, pp.
  627--642, 2016.

\bibitem{DBLP:conf/eccv/LinMBHPRDZ14}
T.~Lin, M.~Maire, S.~J. Belongie, J.~Hays, P.~Perona, D.~Ramanan,
  P.~Doll{\'{a}}r, and C.~L. Zitnick.
\newblock Microsoft {COCO:} common objects in context.
\newblock In {\em 13th European Conference on Computer Vision (ECCV)}, pp.
  740--755, 2014.

\bibitem{Liu:etal:eccv2016:ssd}
W.~Liu, D.~Anguelov, D.~Erhan, C.~Szegedy, S.~E. Reed, C.~Fu, and A.~C. Berg.
\newblock {SSD:} single shot multibox detector.
\newblock In {\em 14th European Conference on Computer Vision (ECCV)}, pp.
  21--37, 2016.

\bibitem{DBLP:conf/iccv/MartinezHRL17}
J.~Martinez, R.~Hossain, J.~Romero, and J.~J. Little.
\newblock A simple yet effective baseline for 3d human pose estimation.
\newblock In {\em IEEE International Conference on Computer Vision (ICCV)}, pp.
  2659--2668, 2017.

\bibitem{Martiez-Gonzalez:etal:iros2018:rpm}
{\'{A}}.~Mart{\'{\i}}nez{-}Gonz{\'{a}}lez, M.~Villamizar, O.~Can{\'{e}}vet, and
  J.~Odobez.
\newblock Real-time convolutional networks for depth-based human pose
  estimation.
\newblock In {\em IEEE/RSJ International Conference on Intelligent Robots and
  Systems (IROS)}, pp. 41--47, 2018.

\bibitem{Mehta:etal:XNECT:TOG2020}
D.~Mehta, O.~Sotnychenko, F.~Mueller, W.~Xu, M.~Elgharib, P.~Fua, H.~Seidel,
  H.~Rhodin, G.~Pons{-}Moll, and C.~Theobalt.
\newblock Xnect: real-time multi-person 3d motion capture with a single {RGB}
  camera.
\newblock {\em {ACM} Trans. Graph.}, 39(4):82, 2020.

\bibitem{Mehta:etal:ACM2017:VNect}
D.~Mehta, S.~Sridhar, O.~Sotnychenko, H.~Rhodin, M.~Shafiei, H.~Seidel, W.~Xu,
  D.~Casas, and C.~Theobalt.
\newblock Vnect: real-time 3d human pose estimation with a single {RGB} camera.
\newblock {\em {ACM} Trans. Graph.}, 36(4):44:1--44:14, 2017.

\bibitem{DBLP:conf/nips/NewellHD17}
A.~Newell, Z.~Huang, and J.~Deng.
\newblock Associative embedding: End-to-end learning for joint detection and
  grouping.
\newblock In I.~Guyon, U.~von Luxburg, S.~Bengio, H.~M. Wallach, R.~Fergus,
  S.~V.~N. Vishwanathan, and R.~Garnett, eds., {\em Annual Conference on Neural
  Information Processing Systems}, pp. 2277--2287, 2017.

\bibitem{Newell:etal:ECCV2016:hourglass}
A.~Newell, K.~Yang, and J.~Deng.
\newblock Stacked hourglass networks for human pose estimation.
\newblock In B.~Leibe, J.~Matas, N.~Sebe, and M.~Welling, eds., {\em 14th
  European Conference on Computer Vision (ECCV)}, pp. 483--499, 2016.

\bibitem{DBLP:conf/cvpr/PapandreouZKTTB17}
G.~Papandreou, T.~Zhu, N.~Kanazawa, A.~Toshev, J.~Tompson, C.~Bregler, and
  K.~Murphy.
\newblock Towards accurate multi-person pose estimation in the wild.
\newblock In {\em IEEE Conference on Computer Vision and Pattern Recognition
  (CVPR)}, 2017.

\bibitem{DBLP:conf/cvpr/PavlakosZD18}
G.~Pavlakos, X.~Zhou, and K.~Daniilidis.
\newblock Ordinal depth supervision for 3d human pose estimation.
\newblock In {\em IEEE Conference on Computer Vision and Pattern Recognition
  (CVPR)}, pp. 7307--7316, 2018.

\bibitem{DBLP:conf/cvpr/PavlakosZDD17a}
G.~Pavlakos, X.~Zhou, K.~G. Derpanis, and K.~Daniilidis.
\newblock Coarse-to-fine volumetric prediction for single-image 3d human pose.
\newblock In {\em IEEE Conference on Computer Vision and Pattern Recognition
  (CVPR)}, pp. 1263--1272, 2017.

\bibitem{Redmon:etal:cvpr16:yolo}
J.~Redmon, S.~K. Divvala, R.~B. Girshick, and A.~Farhadi.
\newblock You only look once: Unified, real-time object detection.
\newblock In {\em IEEE Conference on Computer Vision and Pattern Recognition
  (CVPR)}, pp. 779--788, 2016.

\bibitem{Redmon:Farhadi:cvpr2017:yolo2}
J.~Redmon and A.~Farhadi.
\newblock {YOLO9000:} better, faster, stronger.
\newblock In {\em IEEE Conference on Computer Vision and Pattern Recognition
  (CVPR)}, pp. 6517--6525, 2017.

\bibitem{Ren:etal:pami2017:faster:rcnn}
S.~Ren, K.~He, R.~B. Girshick, and J.~Sun.
\newblock Faster {R-CNN:} towards real-time object detection with region
  proposal networks.
\newblock {\em {IEEE} Trans. Pattern Anal. Mach. Intell.}, 39(6):1137--1149,
  2017.

\bibitem{DBLP:conf/eccv/RhodinRCRST16}
H.~Rhodin, N.~Robertini, D.~Casas, C.~Richardt, H.~Seidel, and C.~Theobalt.
\newblock General automatic human shape and motion capture using volumetric
  contour cues.
\newblock In {\em 14th European Conference on Computer Vision (ECCV)}, pp.
  509--526, 2016.

\bibitem{DBLP:conf/bmvc/TekinKSLF16}
B.~Tekin, I.~Katircioglu, M.~Salzmann, V.~Lepetit, and P.~Fua.
\newblock Structured prediction of 3d human pose with deep neural networks.
\newblock In {\em British Machine Vision Conference (BMVC)}, 2016.

\bibitem{Wang:etal:ACM2016}
K.~Wang, S.~Zhai, H.~Cheng, X.~Liang, and L.~Lin.
\newblock Human pose estimation from depth images via inference embedded
  multi-task learning.
\newblock In {\em {ACM} Conference on Multimedia Conference}, pp. 1227--1236,
  2016.

\bibitem{Wei:etal:CVPR2016:CPM}
S.~Wei, V.~Ramakrishna, T.~Kanade, and Y.~Sheikh.
\newblock Convolutional pose machines.
\newblock In {\em IEEE Conference on Computer Vision and Pattern Recognition
  (CVPR)}, pp. 4724--4732, 2016.

\bibitem{Xiong:etal:iccv2019:a2j}
F.~Xiong, B.~Zhang, Y.~Xiao, Z.~Cao, T.~Yu, J.~T. Zhou, and J.~Yuan.
\newblock {A2J:} anchor-to-joint regression network for 3d articulated pose
  estimation from a single depth image.
\newblock In {\em IEEE/CVF International Conference on Computer Vision (ICCV)},
  pp. 793--802, 2019.

\end{thebibliography}

%%%%%%%%%%%%%%%%%%%%%%%%%%%%%%%%%%%%%%%
%%%%%%%%%%%%%%%%%%%%%%%%%%%%%%%%%%%%%%%
%%%%%%%%%%%%%%%%%%%%%%%%%%%%%%%%%%%%%%%
%%%%%%%%%%%%%%%%%%%%%%%%%%%%%%%%%%%%%%%
%%%%%%%%%%%%%%%%%%%%%%%%%%%%%%%%%%%%%%%

\cleardoublepage

\appendix

\section{Regression networks}
\label{sec:networks}

The network architecture and ground truth preparation for each component are described in detail.

\subsection{Backbone network}
The backbone network is implemented to include layers 0-2 from ResNet-34~\cite{He:etal:CVPR2016:resnet} for general image encoding. It outputs a $\frac{w}{8} \times \frac{h}{8} \times 128$ feature map, where $h$ and $w$ indicate height and width of an input image respectively. We choose to maintain $8\times$ downsampling level in the following functional branches to make part-level inference both efficient and capable of handling human parts at a distance.

\begin{comment}
\begin{figure*}[!h]
    \centering
    \includegraphics[width=0.98\textwidth]{figs/branches_merge.pdf}\\
    (a) Heatmap Branch \qquad\qquad\qquad\quad (b) Depth Branch \qquad\qquad\qquad\quad (c) TPDF Branch
    \caption{Functional branches. (a) The heatmap branch predicts $K$ confidence maps for body parts with an additional map for background. (b) The depth branch outputs $K$ depth maps for $K$ body parts, respectively. (c) The TPDF branch outputs $2K$ maps of displacement vectors $\{X_j\}^{K}_{j=1}$, $\{Y_j\}^{K}_{j=1}$. The field visualization follows the optical flow standard.}
  \label{fig:part:branches}
\end{figure*}
\end{comment}

\subsection{Heatmap branch}

\begin{figure}[!h]
    \centering
    \includegraphics[width=0.4\textwidth]{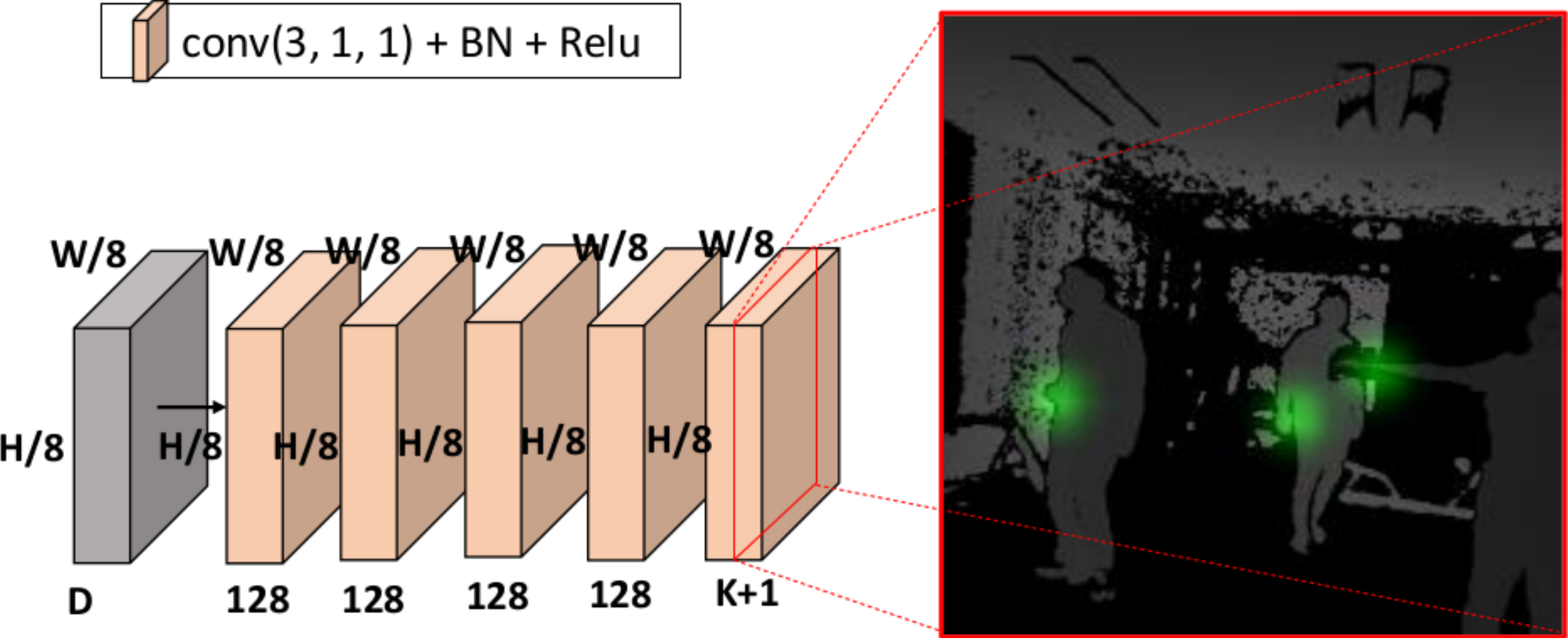}
    \caption{\textbf{Heatmap branch.} The heatmap branch predicts $K$ confidence maps for body parts with an additional map for background. }
  \label{fig:part:ht:branch}
\end{figure}

The heatmap branch is designed to predict $K+1$ confidence maps corresponding to $K$ body part and the background. The heatmap branch is made of five $3\times3$ convolutional layers as illustrated in Figure~\ref{fig:part:ht:branch}. It is worth noting that every convolutional layer mentioned in our method is followed by BN and ReLU layers. To prepare ground-truth part confidence maps $\{H^*_j\}^{K+1}_{j=1}$, we adopt the same method introduced in OpenPose~\cite{Cao:etal:cvpr17:openpose}, which applies a Gaussian filter at each part location.

\subsection{Depth branch}

\begin{figure}[!b]
    \centering
    \includegraphics[width=0.4\textwidth]{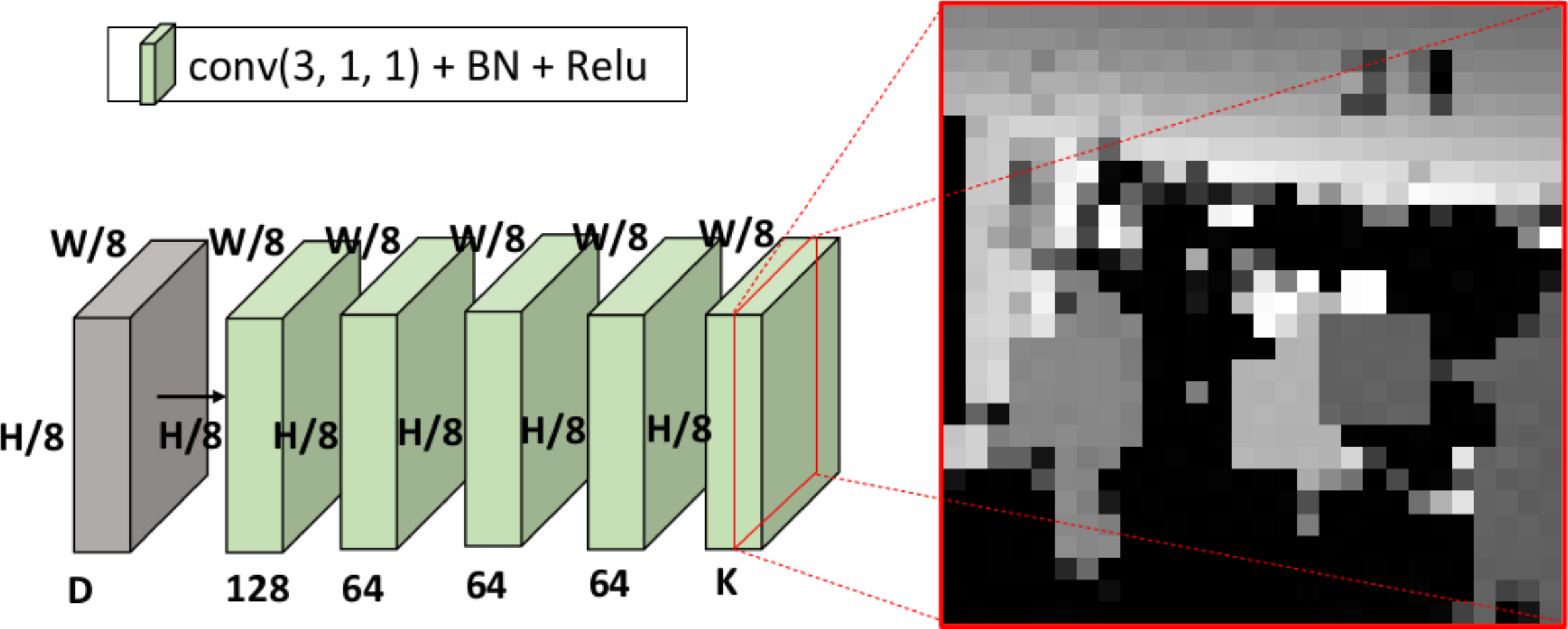}
    \caption{ \textbf{Depth branch.} The depth branch outputs $K$ depth maps for $K$ body parts, respectively.}
  \label{fig:depth:branch}
\end{figure}

The depth branch predicts part-wise depth maps, which is meaningful in relieving the effect from raw depth artifacts and in recovering the true depth of a part under occlusion. The network is made of five convolutional layers whose specific architecture is shown in Figure~\ref{fig:depth:branch}.

To prepare ground-truth depth maps $\{D^*_j\}^K_{j=1}$, each map is initialized with the resized raw depth input. The depth values within a 2-pixel-radius disk centered at each part $j$ are overridden with the ground-truth depth of part $j$, as illustrated in Figure~\ref{fig:depth:branch}. In a multi-person scenario, if a 2D grid position is occupied by masks of more than one part instance, the writing of depth values follows a standard \textit{z}-buffer rule where the smallest depth value is recorded. In addition, the weight maps $\{W^d_j\}^K_{j=1}$ are prepared in the same dimension as the ground-truth depth maps. We use weight $0.9$ for a foreground grid while $0.1$ for the background.

\subsection{TPDF branch}

\begin{figure}[!h]
    \centering
    \includegraphics[width=0.4\textwidth]{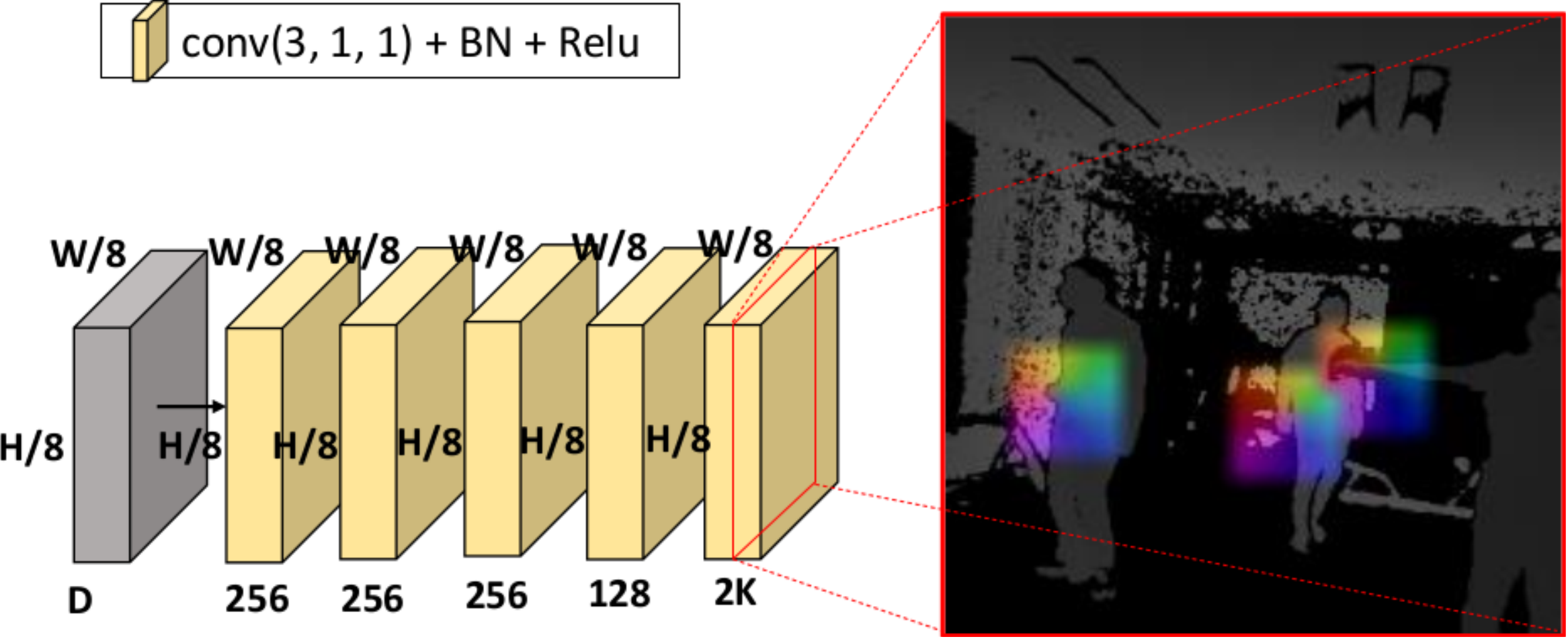}
    \caption{ \textbf{TPDF branch.} The TPDF branch outputs $2K$ maps of displacement vectors $\{X_j\}^{K}_{j=1}$, $\{Y_j\}^{K}_{j=1}$. The field visualization follows the optical flow standard.}
  \label{fig:tpdf:branch}
\end{figure}

TPDF maps are predicted from the TPDF branch implemented following the architecture as shown in Figure~\ref{fig:tpdf:branch}. During ground-truth preparation, $\{X^*_j\}^{K}_{j=1}$ and $\{Y^*_j\}^{K}_{j=1}$ are prepared so that the displacement vector at a 2D position points to the closest part position. Specifically, the displacement vector is only non-zero within the truncated range ($r=2$) from each part position, as shown in Figure~\ref{fig:tpdf:branch}. The preparation of weight maps $\{W^t_j\}^K_{j=1}$ is similar to the process for the depth branch. However, the weights within the truncated mask is set to $1.0$ and the rest is set to strict $0$. %Although the working case for TPDF assumes a body part from a global pose falls in the truncated range, the rest cases are solved by the conflict resolving scheme presented. 

\subsection{Global pose network}

\begin{figure}[!h]
    \centering
    \includegraphics[width=0.45\textwidth]{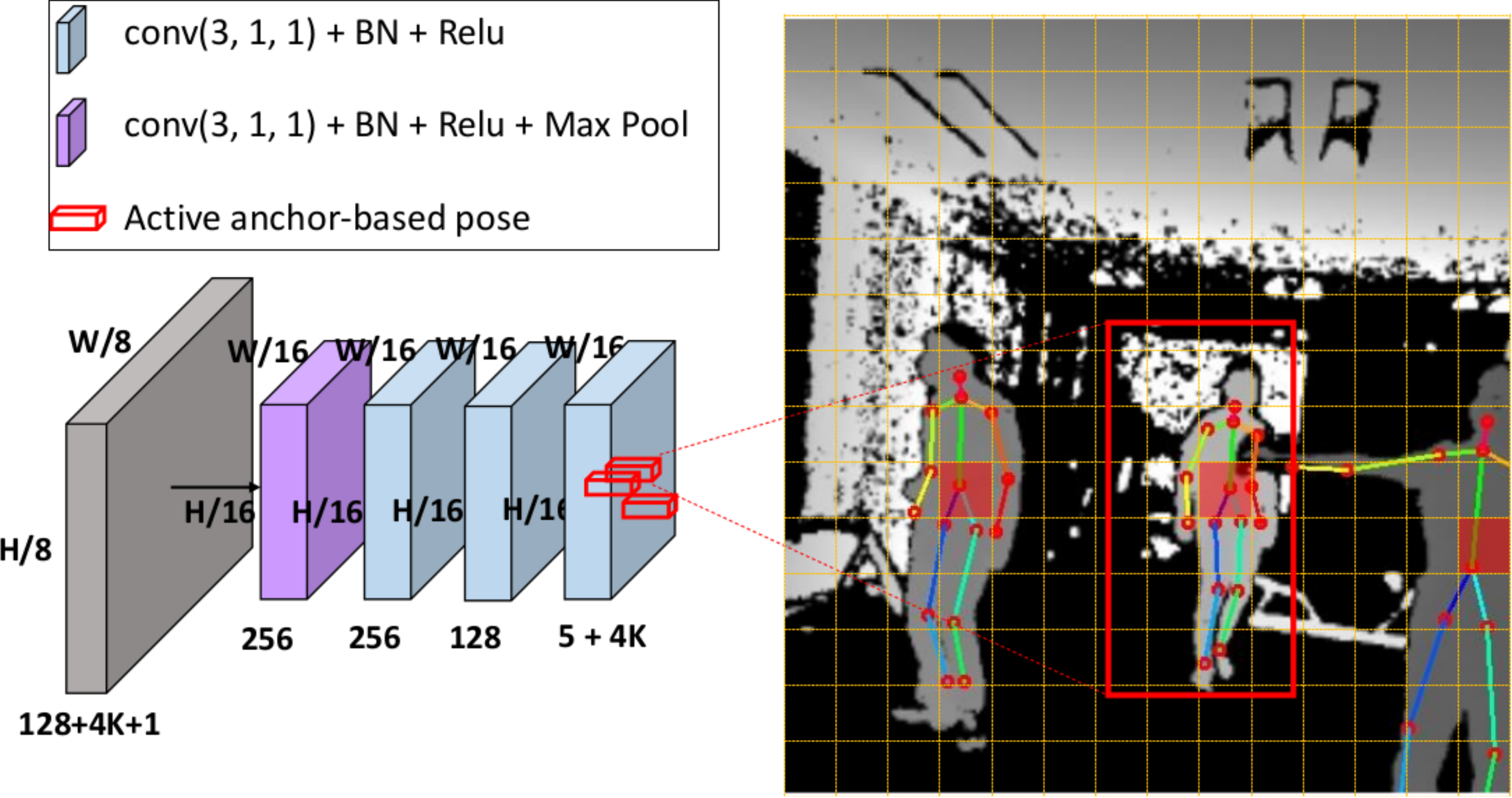}
    \caption{ \textbf{Global pose network}. The global pose network is composed of four $3 \times 3$ convolutional layers, where an additional max-pooling is involved in the first layer. The network outputs an anchor-based global pose map, which is converted to a set of poses after NMS.}
  \label{fig:global:pose:net}
\end{figure}

The global pose network predicts a global pose map from concatenated features from the backbone and functional branches. The network includes four convolutional layers where the first is followed by a max pooling to cast the feature map to $16\times$ downsampling level, as shown in Figure~\ref{fig:global:pose:net}. 

The ground-truth preparation process is similar to Yolo2~\cite{Redmon:Farhadi:cvpr2017:yolo2}. Specifically, the ground-truth global pose map $P^*$ is prepared so that each grid records five bounding box attributes and a set of pose attributes $\{(dx^a_j, dy^a_j, Z^a_j, v^a_j)\}^K_{j=1}$ of the ground-truth pose for each associated anchor $a$. Specifically, $(dx^a_j, dy^a_j)$ indicates the 2D offsets of part $j$ from the anchor center, $Z^a_j$ indicates the 3D part depth, and $v^a_j$ indicates the visibility of part $j$. The value of $v^a_j$ is assigned to $1$ when the depth from a global pose part $Z^a_j$ is different from the corresponding depth branch ground-truth in $D_j$, otherwise it is assigned to $0$. The weight map $W^p$ is prepared in the same dimension as $P^*$. For the dimensions corresponding to bounding box probabilities, $0.9$ is applied to the grids associated with ground truth, while $0.1$ is assigned to the rest. For the other dimensions, the weights are strictly assigned to $1$ or $0$. The weight map is designed in such a  way because the detection task related to $p_b$ considers both foreground and background while the regression task to other attributes focuses only on foreground.

%%%%%%%%%%%%%%%%%%%%%%%%%%%%%%%%%%%%%%%%%%%%%%%%%%%%%%%%%%%%%%%%%%%%%%%%%%%%%%%%%%%%%%%%%%%%%%
%%%%%%%%%%%%%%%%%%%%%%%%%%%%%%%%%%%%%%%%%%%%%%%%%%%%%%%%%%%%%%%%%%%%%%%%%%%%%%%%%%%%%%%%%%%%%%
%%%%%%%%%%%%%%%%%%%%%%%%%%%%%%%%%%%%%%%%%%%%%%%%%%%%%%%%%%%%%%%%%%%%%%%%%%%%%%%%%%%%%%%%%%%%%%

\section{Depth Augmentation}
\label{sec:depth:aug}

Given camera intrinsic parameters, the captured depth map, and associated 2D/3D poses, novel depth maps and associated 2D/3D poses can be generated via simulating the camera re-positioned along the principle axis. Specifically, suppose a 3D point $(X,Y, Z_0)$ in the original camera coordinate frame with projection at $(x_0, y_0)$ in the original image is placed at $(X,Y, Z_1)$ in the new camera coordinate frame and be projected to $(x_1, y_1)$ in the new image, we can write the following relationships based on similar triangles:

\begin{align}
\frac{X}{x_1 - cx} = \frac{Z_1}{f} = \frac{Y}{y_1 - cy}\\
\frac{X}{x_0 - cx} = \frac{Z_0}{f} = \frac{Y}{y_0 - cy}
\end{align}
where $(cx, cy)$ represents the principle point in both images, and $f$ indicates the focal length. Dividing the first equation by the second, we get:
\begin{equation}
    a = \frac{x_0 -  cx}{x_1 - cx} = \frac{y_0 -  cy}{y_1 - cy} = \frac{Z_1}{Z_0}
\end{equation}
Thus, a new depth image can be simply generated via randomly sampling $a$ within a reasonable range, and mapping the area defined by the original four image corners to the new locations in the new image. Meanwhile, the depth values of the new image and the associated 2D and 3D body part positions can also be calculated. This depth augmentation method is rather effective. However, it can not simulate the dis-occlusion from a different camera location, such that the augmented depth data can not fully represent the quality of real captured data. In practice, the synthesized depth is directly determined by the original depth and $a$, whose effect is analysed in Section~\ref{sec:detail:ablation}.

\section{Multi-Person Data Augmentation}
\label{sec:mpaug}

\begin{figure}[!h]
    \centering
    \includegraphics[width=0.18\textwidth]{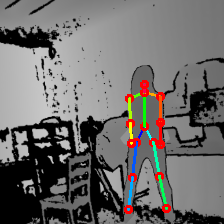}
    \includegraphics[width=0.18\textwidth]{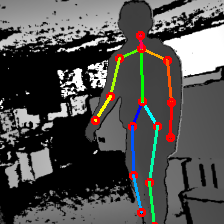}\\
    \includegraphics[width=0.18\textwidth]{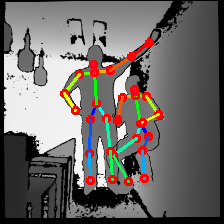}
    \includegraphics[width=0.18\textwidth]{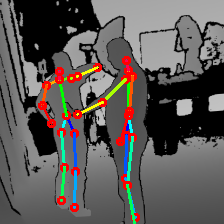}
  \caption{\textbf{Augmented training samples.} (Top) Single-person training samples augmented with a random background scene. (Bottom) Augmented multi-person training sample composed from multiple single-person training samples and a random background scene.}
  \label{fig:data:aug}
\end{figure}

Multi-person and background data augmentation plays an important role in training models generalizable to uncontrolled multi-person scenarios. Such augmentation is enabled by the training data of MP-3DHP, which not only includes ground-truth 3D joint positions but also the foreground masks. Specifically, the training set includes human subjects recorded at four different locations relative to the camera plus a set of free-style movements, as shown in Figure~\ref{fig:MP-3DHP:dataset} (top) and a set of background-only images as shown in the left two images in Figure~\ref{fig:MP-3DHP:dataset} (bottom). Given a set of background-only images, a human segment from the training set can be used to simply override the pixels within the same region, leading to a background augmented image as shown in Figure~\ref{fig:data:aug} (Top). Similarly, human segments from different recording locations can be composed with random background images following the \textit{z}-buffer rule to generate multi-person augmented images as shown in Figure~\ref{fig:data:aug} (Bottom). 

There are a few heuristics associated with the simple augmentation. First, we include no more than two bodies in the multi-person augmentation with an assumption that inter-person occlusion cases between two bodies can well represent the inter-person occlusion cases between more bodies. Second, the straight-forward composition does not consider scene geometry, thus some generated cases appear unrealistic. However, the conflict with scene geometry is not considered a serious issue in training because all the pipelines only adopt convolutional layers learning that only relies on the local context between a body part and the background in its vicinity rather than the whole scene. Finally, there are sensor artifacts around each human segment that can not be perfectly removed. This issue indeed affects the generalization capability of a trained model to the real data. For example, an occluded part from the augmented data is still roughly visible because of the black margin around the human segment, however an occluded part appears truly invisible in real data. Examples of multi-person augmentation and background augmentation are visualized in Figure~\ref{fig:data:aug}.

\section{Application}

\begin{figure}[!h]
    \centering
    \includegraphics[width=0.4\textwidth]{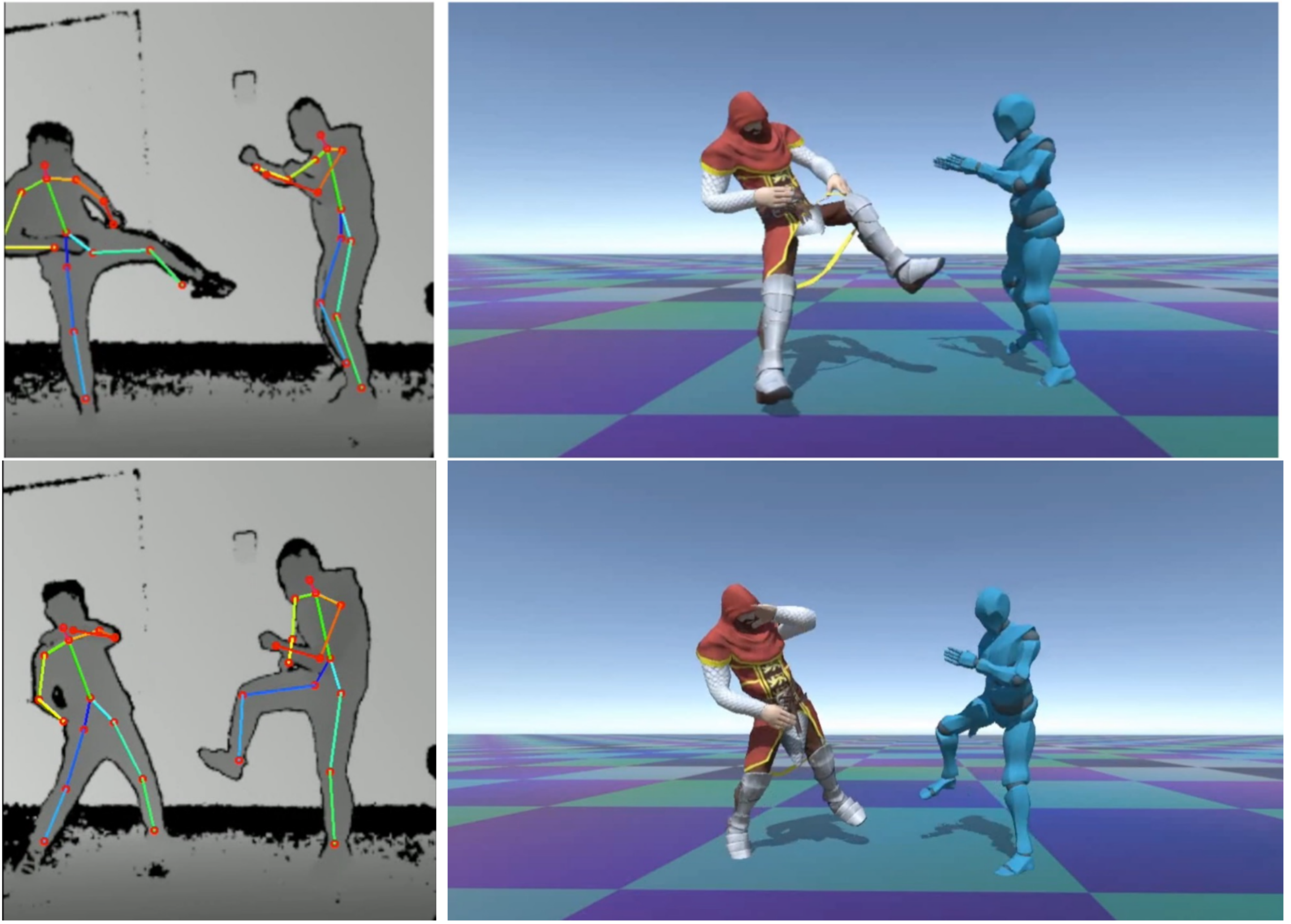}
    \caption{ \textbf{Virtual avatar driving results.} The left column shows the input depth images and the right column shows the corresponding virtual avatar interaction.}
  \label{fig:avatar_fig}
\end{figure}

For AR/VR applications, we demonstrate that our prediction of 3D human body parts enables the virtual avatar driving where 3D motion capture plays a key role. As shown in~\autoref{fig:avatar_fig}, we convert a sequence of predicted 3D joint positions into the rotation angles of each joint to drive the animated virtual avatar. The supplemental video shows a frame-by-frame avatar-driving animation, and the result is further smoothed by inter-frame filtering. Here, we show the result by recovering the rotation angle of each joint, and the pelvis position is fixed in a certain location.

\end{document}